\documentclass[11pt,a4paper]{article}
\usepackage{emnlp2020}

\usepackage{times}
\usepackage{latexsym}

\usepackage{graphicx}
\usepackage{caption}
\usepackage{subcaption}
\usepackage{amsmath}

\usepackage{multirow}
\usepackage{caption}
\usepackage{subcaption}
\usepackage[inline]{enumitem}
\usepackage{varwidth}
\usepackage{todonotes}
\usepackage[framemethod=TikZ]{mdframed}
\usepackage{amsfonts}
\usepackage{array,booktabs,arydshln,xcolor}
\usepackage{multirow}
\usepackage{graphics}
 \usepackage[most]{tcolorbox}
\usepackage{lipsum}
\usepackage{pifont}
\usepackage{caption}
\usepackage{subcaption}
\usepackage{hyperref}
\usepackage{makecell}

\usepackage{microtype}

\aclfinalcopy 

\usepackage{algorithmicx}
 \usepackage{algpseudocode}

\title{Weakly Supervised Learning of Nuanced Frames for Analyzing Polarization in News Media}

\author{Shamik Roy\\
  Department of Computer Science\\
  Purdue University\\
  West Lafayette, IN 47907\\
  \texttt{roy98@purdue.edu} \\\And
  Dan Goldwasser \\
  Department of Computer Science\\
  Purdue University\\
  West Lafayette, IN 47907\\
  \texttt{dgoldwas@purdue.edu} \\
  }

\date{}

\begin{document}
\maketitle
\begin{abstract}
In this paper we suggest a minimally-supervised approach for identifying nuanced frames in news article coverage of politically divisive topics. We suggest to break the broad policy frames suggested by~\citeauthor{boydstun2014tracking}, \citeyear{boydstun2014tracking} into fine-grained subframes which can capture differences in political ideology in a better way. We evaluate the suggested subframes and their embedding, learned using minimal supervision, over three topics, namely, immigration, gun-control and abortion. We demonstrate the ability of the subframes to capture ideological differences and analyze political discourse in news media.


\end{abstract}

\section{Introduction}\label{sec:introduction}

As the political climate and the news media in the United States become increasingly polarized~\cite{prior2013media,pew2018}, it is important to understand the perspectives underlying the political divisions and analyze their differences. 
Political framing, studied by political scientists~\cite{entman1993framing,chong2007framing}, provides the means to study these perspectives. It is a nuanced political strategy, used to bias the discussion on an issue towards a specific stance by emphasizing specific aspects that prime the reader to accept that stance.  
%
To help clarify this definition, consider two articles on the highly polarized immigration issue.\\

\noindent\textbf{Example 1:}\textit{ Different Perspectives on Immigration}
 
\begin{tcbraster}[raster columns=1,raster equal height=rows,raster valign=top, size=small]
\begin{tcolorbox}[colback=blue!15!white,colframe=blue!75!black,nobeforeafter, title={ \textit{Adapted from} \bf{Alternet }}~~(\textit{Left})]
  \small{{
%
Employees-many of whom are undocumented immigrants from Mexico, Ecuador and elsewhere-toil seven days a week for less than minimum wage, with no overtime pay.
}}
\end{tcolorbox}  
\begin{tcolorbox}[colback=red!15!white,colframe=red!75!black,nobeforeafter, title={ \textit{Adapted from} \bf{Breitbart }}~~(\textit{Right})]
  \small{{
Mass immigration has come at the expense of America's working and middle
class, which suffered from poor job growth, stagnant wages, and increased public costs.
  }}
\end{tcolorbox}  
\end{tcbraster}
\vspace{7pt}

The two articles capture opposite political perspectives, liberal (top) and conservative (bottom). They do not directly contradict each other, instead they focus the discussion on different aspects helping them argue their case. The first emphasizing the deprivation of minimum wage for immigrants, and the second emphasizing implication on wages for U.S. workers. This process is known as \textit{framing}. 
Our goal is to define, and automatically identify, relevant framing dimensions in politically-motivated coverage of news events \textit{to the extent they can capture and explain the differences in perspectives} across the conservative-liberal ideological divide~\cite{ellis2012ideology,preoctiuc2017beyond}. We focus on three divisive topics --  immigration, gun-control and abortion. 

  Previous work by~\citet{boydstun2014tracking} studied policy issue framing on news media and suggested 15 broad frames to analyze how issues are framed, which include \textit{economic, morality} and \textit{security}, among others. These framing dimensions can help capture ideological splits~\cite{johnson2017ideological}. For example, by framing the immigration issue using the morality frame or using the security frame, the reader is primed to accept the liberal or conservative perspectives, respectively. However, as shown in Example 1, in some cases this analysis is too coarse grained, as both articles frame the issue using the economic frame, suggesting that a finer grained analysis is needed to capture the differences in perspective. To help resolve this issue, we suggest a data-driven refinement, trained with minimal supervision effort.
  
  Our approach works in three steps. First, we construct topic-specific lexicons capturing the way the frames are instantiated in each topic. 
  In the second step, we identify repeating expressions used in the context of the different frames, and group them to form \textit{subframes} which separate between different usages of the same frame to express different political perspectives. 
  In Example 1, \textit{Minimum Wage Economy} and \textit{Salary Stagnation} are both subframes of the Economy frame, which capture the ideological differences in the two articles.
  We use external knowledge sources to identify relevant subframes for each topic and rely on human judgements to match the repeating expressions with these subframes. 
  Finally, we exploit this resource to train an embedding model, which represents in the same space the subframes labels, the lexicon containing subframes indicator expressions and paragraphs extracted from news articles containing these expressions. The embedding model captures the context in which subframe appear, and as a result can generalize and capture subframe usage in new texts.
   
  Our approach can be viewed as a middle ground between event-specific frames, emerging from the data and capturing properties unique to the given topic~\cite{tsur2015frame,demszky2019analyzing}, and general issue frames~\cite{boydstun2014tracking,card-EtAl:2015:ACL-IJCNLP,johnson2017leveraging,field2018framing,hartmann2019issue} that use the same set of framing dimensions for all topics. On the one hand, it can capture nuanced, topic-specific subframes, while on the other, it maps these subframes into general framing dimensions. In the above example, it allows us to identify that the economy frame is important for both the liberal and conservative perspectives on immigration, despite the fact that it is instantiated using a different subframe.
   
 We evaluate the quality of the learned model in several ways, by applying it politically-motivated news article coverage of divisive topics. First, we show that the lexicon we developed and the induced sub-frames can effectively separate between ideological standpoints expressed in the articles. Second, we evaluate the quality of the learned model, showing that subframe labels assigned to new paragraphs correlate well with human judgements. Finally, we use the model to analyze the different perspectives in left and right leaning news coverage, and their change over time.
%





\section{Related Work}\label{sec:related}
Understanding and analyzing political perspectives in news coverage
has gathered significant interest in recent years~\cite{lin2006side,greene2009more,iyyer2014political,li2019encoding,fan2019plain,jiang2019,hanawa2019sally}, broadly related to analysis of bias or partisanship and expressions of implicit sentiment~\cite{recasens2013linguistic,baumer2015testing,field2018framing,gentzkow2016measuring, monroe2008fightin,an2019political,menini2017topic}.
In addition to predicting the underlying perspective, our work focuses on explaining the perspectives underlying the ideological coverage of news events. We build specifically on issue-frames~\cite{boydstun2014tracking}, however our work is related to framing and agenda setting analysis work more broadly~\cite{tsur2015frame,baumer2015testing,fulgoni-etal-2016-empirical,field2018framing,demszky2019analyzing}.

\section{Data Collection }
 We collected $21,645$ news articles on three politically polarized topics, \textit{abortion, immigration} and \textit{gun control}. We used the hyper-partisan news dataset \cite{kiesel2019semeval} and we crawled additional news articles on the topics from sources with known political bias provided by \url{mediabiasfactcheck.com}, where the articles are categorized based on their topics on the websites of the sources. All the news articles are U.S. politics based and written in English. We identify the topic of news articles in the hyper-partisan news dataset by looking at presence of certain keywords in the titles and urls of news articles. For example, `abortion' for topic \textit{abortion}; `migrant', `migration' for topic \textit{immigration}; `gun' for topic \textit{gun control}. In case of absence of any of the keywords in the title or url, we annotate the article with the corresponding topic if the keywords appear at least $3$ times in the article text. The hyper-partisan news dataset provides bias-labels of the news articles and we labeled our crawled news articles based on their source bias according to \url{mediabiasfactcheck.com}. We consider only the left and right biased news articles. The dataset is summarized in Table \ref{tab:dataset-summary}.
 
\begin{table}[ht]
\centering
\scalebox{0.69}{
\begin{tabular}{>{\arraybackslash}m{3.3cm}|>{\centering\arraybackslash}m{1.7cm}|>{\centering\arraybackslash}m{2cm}|>{\centering\arraybackslash}m{2.1cm}}
 \hline
 & \textbf{Abortion} & \textbf{Immigration} & \textbf{Gun Control} \\
 \hline
 \# of News Articles & 6,476 & 8,516 & 6,653\\
 \# of Left Articles & 3,437 & 3,496 & 3,198\\
 \# of Right Articles & 3,039 & 5,020 & 3,455\\
 \# of Paragraphs & 106,931 & 135,479 & 95,872\\
 Span of Year & 1984-2019 & 2000-2019 & 1996-2019\\
 $>$80$\%$-Articles Since & 2010 & 2016 & 2011\\
 \hline
\end{tabular}}
\caption{Dataset Summary. (Articles are split into paragraphs by newline character.)}
\label{tab:dataset-summary}
\end{table}
\section{Modeling Political Framing}
Our goal in this paper is to identify framing dimensions that can be used to capture difference between the two ideological polarities. 
%
The frames used for this analysis could be issue specific, or as suggested by \citet{boydstun2014tracking}, generalize over several policy issues, using a fixed set of framing dimensions. In many cases, as we show in this paper, the general policy frames do not capture the nuanced aspects of the issue highlighted by each side to bias the discussion. In Example 1, \textbf{both} sides use the \textit{economy} frame, however it is instantiated in different ways, to promote opposite views.  To help combat this issue, we suggest a middle ground between generalized policy frames and event-specific frames, by constructing a hierarchy of frames and sub-frames, the first derived from the definitions and data of media frames corpus~\cite{card2015media}, and the second emerging from the data directly, by tracking the differences in the vocabulary used when these frames are instantiated in different topics, and grouping them to sub-frames.

This process takes place in three steps. First, we create a lexicon of topic specific phrases capturing how the frame is invoked in each policy issue. Second, we manually group these phrases into subframes. Finally, we embed the sub-frames using weak-supervision, allowing us to associate subframes with new text, beyond the extracted phrases. The following subsections explain each step.

\subsection{Extending Frame Lexicon}\label{sec:lex}
\paragraph{Step 1: Annotate news article paragraphs with policy frames.}We follow the procedure suggested and validated by \citet{field2018framing}, and use the media frame corpus to derive a unigram lexicon for each of the $15$ policy frames (\citet{boydstun2014tracking}) based on their Pointwise Mutual Information (PMI) \cite{church1990word}. We use the $250$ top-PMI words for each of the frames. We discard unigrams which occur in less than $0.5\%$ of the documents and more than $98\%$ of the documents, the same thresholds used by \citet{field2018framing}. 
We define these unigrams as \textit{\textbf{frame indicators}}, and use them to annotate our data.  We break the articles in our dataset into paragraphs, and  annotate them by ranking frames using the number of lexicon matches. We use the top $2$ frames per paragraph. Since news articles can cover a topic from multiple angles, we identify frames in paragraph level instead of article level.
\\
\paragraph{Step 2: Building Topic-specific Lexicons.}We hypothesize that the frame-level analysis cannot capture nuanced talking points. In Example 1, both texts use the \textit{Economic} frame using same unigram indicator (`wage'). However, other words in the text can help identify the nuanced talking points (e.g., `minimum wage' in case of left and `stagnant wages' in case of right). We follow this intuition, and extend the frame-level dictionary to topic-specific frame lexicon, using bi-gram and tri-gram phrases extracted from the annotated paragraphs with frames in Step 1. For an n-gram $g$ we calculate the PMI with frame $f$, $I(g, f)$, as follows:
\begin{align*}
     I(g,f)=\operatorname{log}\frac{P(g|f)}{P(g)}
\end{align*}{}
Where $P(g|f)$ is computed by taking all paragraphs annotated with frame $f$ and computing $\frac{count(g)}{count(allngrams)}$ and similarly, $P(g)$ is computed by counting n-gram $g$ over the whole corpus. We assign each n-gram to the frame with highest PMI score and build an n-gram lexicon for each frame. We did not consider bi-grams or tri-grams appearing in more than $50\%$ of the paragraphs and less than $0.02\%$ of the paragraphs. The process is topic-specific, resulting in three lexicons, one for each topic we study in this paper. Following this procedure we found $4,116$ bigrams and $1,787$ trigrams for the topic \textit{abortion}, $3,293$ bigrams and $1,451$ trigrams for the topic \textit{gun control}, $3,743$ bigrams and $1,385$ trigrams for the topic \textit{immigration}. We define these n-grams as \textit{\textbf{subframe indicators}}.

\paragraph{Step 3: Lexicon Validation.}
We hypothesize that the topic-specific subframe indicators capture political perspective better than the frame indicators. 
To validate this claim we compare the correlation between the usage of frame and subframe indicators in left and right biased news articles. 
We break the dataset into left and right biased documents. Each group is associated with a ranked list of frames and sub-frames indicators, based on their averaged tf-idf scores in all documents with the same political bias.
Then we compare the ranks using Spearman's Rank Correlation Coefficient \cite{zar2005spearman} where coefficient $1$ means perfect correlation. Table \ref{tab:polarization} validates our claim. It shows that the frame indicator lists have a much higher correlation compared to subframe indicators, indicating their usage by both sides with similar importance. 

We investigate the expressivity of the two indicator groups by using them as one-hot features when classifying the political bias of documents using a simple logistic regression classifier. The results in Table \ref{tab:classification-results} shows that the subframe indicators are better in all of the three topics. We also included two strong contextualized feature representation: BERT (base-uncased) \cite{devlin2018bert} and hierarchical-LSTM (HLSTM) \cite{hochreiter1997long}. BERT used the first $512$ tokens of each  article. In HLSTM, we run a biLSTM over the $300d$ GLOVE \cite{pennington2014glove} words embeddings in each paragraph, and average their hidden states to repreent the paragraph, a second bidirectional LSTM was used to create the final representation for the news article in $300d$. The subframe indicator feature outperforms HLSTM in all three topics and BERT in case of immigration.

\begin{table}[t]
\begin{center}
\scalebox{0.67}{
\begin{tabular}{>{\arraybackslash}m{2cm}|>{\centering\arraybackslash}m{4cm}|>{\centering\arraybackslash}m{4cm}}
 \hline
 \textsc{Topics} & \textsc{Frame Indicators Rank Corr. Coef.} & \textsc{Subframe Indicators Rank Corr. Coef.} \\
 \hline
 Abortion       &   0.94 (0.017)    &   0.35 (0.128)\\
 Immigration    &   0.91 (0.018)    &   0.25 (0.142)\\
 Gun Control    &   0.94 (0.011)    &   0.40 (0.112)\\
 \hline
\end{tabular}}
\caption{Average rank correlation coefficient of frame and subframe indicators' ranks between left and right over $15$ policy frames with standard deviations in the brackets. Correlations in individual frames can be found in Appendix A.}
\label{tab:polarization}
\end{center}
\end{table}

\begin{table}[t]
\begin{center}
\scalebox{0.71}{
\begin{tabular}{>{\arraybackslash}m{4cm}|>{\centering\arraybackslash}m{1.7cm}|>{\centering\arraybackslash}m{1.7cm}|>{\centering\arraybackslash}m{1.7cm}}
 \hline
 \textsc{Models} & \textsc{Abort.} & \textsc{Imm.} & \textsc{Gun} \\
 \hline
 LR (Frame Indicators) & 74.57 (0.6) & 82.36 (0.6) & 70.62 (0.8)\\
 LR (Subframe Indicators) & 81.47 (0.2) & 85.31 (0.2)  & 72.34 (0.5) \\
 BERT & 81.58 (1.8) & 79.72 (3.7) & 73.21 (0.7)\\
 HLSTM & 81.12 (0.4) & 84.69 (1.7) & 71.08 (2.8) \\
 \hline
\end{tabular}}
\caption{Test F1 scores (standard deviation) of article classification task (left/right) using 3-fold CV. LR is for Logistic Regression with type of feature in the bracket.}
\label{tab:classification-results}
\end{center}
\end{table}

\subsection{Identification of Subframes} 
Our next step is to identify the nuanced subframes captured by the subframe indicators lexicon we extracted. We use human knowledge to guide this process, such that each general frame can be decomposed into multiple sub-frames, by 
 grouping repeating subframe indicators (i.e. n-grams associated with each frame) to known political talking points. 
 For example, in case of \textit{abortion}, the phrases `Hobby Lobby', `freedom restoration act' discuss a similar issue and can be grouped together to form a subframe (which we denoted as `Hobby Lobby'). We extracted the talking points from Wikipedia and \url{ontheissues.com}, which maintains political perspectives on these issues. We did not consider the frames `Political' and `Others' categories, as all of the three topics are political. We only focused on frames relevant for our three topics. For example,  `Security and Defense' is not related to the topic \textit{abortion}.  Table \ref{tab:subframe-description}, shows all the identified subframes, their parent frames. The subframes' full definition and associated n-grams can be found in Appendix C and B. 

\begin{table}[ht]
\begin{center}
 \scalebox{0.58}{\begin{tabular}{|>{\arraybackslash}m{4cm}|>{\arraybackslash}m{4cm}|>{\arraybackslash}m{4cm}|} 
 \hline 
  \textbf{\textsc{Abortion}} &    \textbf{\textsc{Immigration}}   &   \textbf{\textsc{Gun Control}}\\
 \hline
 
 \makecell[tl]{\textbf{Economic:}\\- Health Care\\- Abort. Provider\\Economy\\- Abortion Funding\\\textbf{Fairness \& Equality:}\\- Reproduction Right\\- Right of Human Life\\\textbf{Legality, Constitution-}\\\textbf{ality, Jurisdiction:}\\- Hobby Lobby\\- Late Term Abortion\\- Roe V. Wade\\\textbf{Crime \& Punishment:}\\- Stem Cell Research\\- Sale of Fetal Tissue\\- Sexual Assault Victims\\\textbf{Health \& Safety:}\\- Birth Control\\\textbf{Morality:}\\- Sanctity of Life\\- Women Freedom\\\textbf{Quality of Life:}\\- Planned Parenthood\\- Pregnancy Centers\\- Life protection\\\textbf{Public Sentiment:}\\- Pro-Life\\- Anti-Abortion\\- Pro-Choice\\\\} 
 & 
 \makecell[tl]{\textbf{Economic:}\\- Minimum Wage\\- Salary Stagnation\\- Wealth Gap\\- Cheap labor availability\\- Taxpayer Money\\\textbf{Crime \& Punishment}\\- Deportation: Illegal\\Immigrants\\- Deportation: In General\\- Detention\\\textbf{Security \& Defense}\\- Terrorism\\- Border Protection\\\textbf{Legality, Const., Juri}\\- Asylum\\- Refugee\\- Birth citizenship \&\\14th Amendment\\\textbf{Policy Pres. \& Eval.}\\- Amnesty\\- Dream Act\\- Family Separation Policy\\- DACA\\\textbf{Fairness \& Equality}\\- Racism \& Xenophobia\\- Merit Based Immigration\\- Human Right\\\textbf{Cultural Identity}\\- Racial Identity\\- Born identity}
 &
 \makecell[l]{\textbf{Economic}\\- Gun Buyback Program\\- Gun Business\\\textbf{Capacity \& Resource}\\- School Safety\\\textbf{Cultural Identity}\\- White Identity\\- Person of Color Identity\\\textbf{Legality, Constitution-}\\\textbf{ality, Jurisdiction:}\\- Ban on Handgun\\- Second Amendment\\- Concealed Carry\\Reciprocity Act\\- Gun Control to\\Restrain Violence\\\textbf{Crime \& Punishment}\\- Illegal Gun\\- Gun Show Loophole\\\textbf{Security \& Defense}\\- Background Check\\- Terrorist Attack\\\textbf{Health \& Safety}\\- Gun Research\\- Mental Health\\- Gun Homicide\\\textbf{Policy Pres. \& Eval.}\\- Assault Weapon\\\textbf{Morality}\\- Right to Self-Defense\\- Stop Gun Crime}\\
 \hline 

 \end{tabular}}
\caption{Subframes with corresponding frames.}
\label{tab:subframe-description}
\end{center}
\end{table}



\subsection{Weakly Supervised Categorization of Subframes}\label{sec:embedding}
In the previous steps we identified relevant sub-frames for each issue and mapped them to the appropriate topic-specific indicators. The indicators can be used for annotating the text directly, as suggested by~\cite{field2018framing}, however we note that they only cover  $16.03\%$, $11.51\%$ and $11.22\%$ of the paragraphs in the topics abortion, immigration and gun control respectively. Instead we use the indicators as a seed set for a weakly-supervised learning process, which intends to generalize the subframe analysis to new text that does not contain the seed subframe indicators, by capturing the relevant context in which these indicators appear.


To identify subframes in paragraphs that do not contain a subframe indicator, we embed the news articles, broken into paragraphs,  the complete subframe indicator lexicon and the subframe labels in a common embedding space. The embedding space is shaped by following  two objectives: (1) the similarity between the embedding of a paragraph and a subframe indicator is maximized if it appears in the paragraph, (2) the similarity between the embedding of a subframe indicator and its corresponding subframe is maximized. 
We briefly describe the embedding learning objective as follows. Given an instance $o$, a positive example $m^{p}$ and a negative example $m^{n}$, where $o$ is needed to be closer to $m^{p}$ and far from $m^{n}$ in the embedding space, the embedding loss is defined:
\begin{align*}
    E_{r}(o,m^{p},m^{n})=l(\operatorname{sim}(o, m^{p}), \operatorname{sim}(o, m^{n}))
\end{align*}
Here, $E_{r}$ defines the embedding loss for objective type $r$ (paragraph to subframe indicator or subframe indicator to subframe label). Our goal is to maximize the similarity of a node embedding with a positive example and minimize the similarity with a negative example. We call a subframe indicator a positive example for a paragraph, if the paragraph contains the subframe indicator. Similarly, a subframe label is a positive example for a subframe indicator, if it is an indicator of the subframe. We randomly sample $5$ negative examples for each positive example from the subset of subframe indicator not present in a paragraph and do the same in case of the subframe indicator to subframe label objective. As, a similarity function ($sim()$) we use dot product and $l()$ is cross-entropy loss which is defined as follow.
\begin{align*}
    l(p,n)=-\operatorname{log}(\frac{e^{\operatorname{sim}(o, m^{p})}}{e^{\operatorname{sim}(o,m^{p})}+e^{\operatorname{sim}(o,m^{n})}})
\end{align*}
Now, for all kind of objectives we can minimize the summed loss $\sum_{r \in R} \lambda_{r} E_{r}$, where $R$ is the set of all kind of objective functions and $\lambda_{r}$ is the weight of loss for objective function of type $r$. We initialized $\lambda_{r}=1$, for both objectives.

We initialize the embeddings of subframe indicators and subframe labels randomly and the paragraph embeddings are obtained by running a bidirectional-LSTM \cite{hochreiter1997long} over the Glove \cite{pennington2014glove} word embeddings of the words of the paragraph. We concatenated the hidden states of the two opposite directional LSTMs to get representation over one time-stamp and average the representations of all time-stamps to get a single representation of the paragraph. All the embeddings are initialized in a $300d$ space. We train this bidirectional-LSTM jointly with the embedding learning. We stop learning if the embeddings learning loss does not decrease for $10$ epochs or reach at $100$ epochs. Dataset and codes can be found at \url{https://github.com/ShamikRoy/Subframe-Prediction}

After the embeddings learning we can get a distribution over all of the subframe labels for each paragraph which is based on the cosine similarity between the embeddings of the paragraph and subframe labels. Thus our model combined with the labeled n-grams have the ability to expand the subframe labels to unlabeled text from other domains of the same topics without any human evaluation.

\paragraph{Evaluating the Embedding Space.} We evaluate the resulting embedding in two ways. First, we interpret the correctness of the \textit{subframe representation} in the embedding space by evaluating whether the paragraphs most similar to it, actually express that subframe based on human judgement. In the second evaluation, we randomly sample articles and use the embedding space similarity to predict relevant subframes. In both cases we intentionally use instances that \textbf{do not} include explicit subframe indicators, in order to evaluate the model's ability to generalize beyond the lexicon. We compare our embedding space to topic-model baseline, using the same subframe indicators as a seed set.

\paragraph{Topic Model Baseline.} We compare our model with guided LDA \cite{jagarlamudi2012incorporating}, a variant of LDA~\cite{blei2003latent} where topics can be guided based on world knowledge. Traditional LDA assigns uniform Dirichlet prior to each word over all topics. Guided LDA assigns more bias to the seed words of a topic which are believed to be true representatives of the topic. Considering each subframe as a topic, we used the annotated indicators as the seed phrases. We learn the guided LDA model over stemmed unigrams, bigrams and trigrams. To omit out stop words and rarely used words we discarded phrases occurring in less than $0.005\%$ and more than $80\%$ of the paragraphs. Now, we compare our model with guided LDA in the following two ways.

\textbf{(1) Subframe Prediction Evaluation.} We evaluate our model's performance when identifying subframes in new paragraphs compared to guided LDA. We take top $20$ most similar paragraphs under each subframe identified by each model and ask human annotators to evaluate if the content of the paragraphs match with the subframe label. We chose the paragraphs which did not have the seed n-grams in their content, so we take the paragraphs that are newly identified. For abortion, immigration and gun control, it resulted in $400$, $440$ and $380$ such examples respectively. We shuffled the paragraphs identified by the two models while presenting to the annotators. We asked $2$ graduate students individually to perform the task by providing them with the subframe descriptions (in Appendix C). We found the Cohen's kappa \cite{cohen1960coefficient} score between the annotators to be $0.83$ implying almost perfect agreement. In case of a disagreement, we asked a 3rd annotator to break the tie who is a researcher in Computational Social Science. Based on the majority voting our model outperforms guided LDA (Table \ref{tab:paragraph-classification-results}).

\begin{table}[ht]
\begin{center}
\scalebox{0.71}{
\begin{tabular}{>{\arraybackslash}m{4cm}|>{\centering\arraybackslash}m{1.7cm}|>{\centering\arraybackslash}m{1.7cm}|>{\centering\arraybackslash}m{1.7cm}}
 \hline
 \textsc{Models} & \textsc{Abort.} & \textsc{Imm.} & \textsc{Gun} \\
 \hline
 Guided LDA & 42.00\% & 39.77\% & 42.63\%\\
 Our Model & 95.25\% & 87.01\% & 90.53\%\\
 \hline
\end{tabular}}
\caption{\% of paragraphs with matching subframe out of $400$, $440$ and $380$ examples per model for the topics Abortion, Immigration and Gun Control respectively.}
\label{tab:paragraph-classification-results}
\end{center}
\end{table}

\textbf{(2) Identifying talking points in news articles.} We can identify the main talking points of a news article based on the distribution of subframes for each paragraph in the news article. We take the average distribution of all the paragraphs and output top-$k$ most probable subframes as summary of the news article. To reduce noise,  we restrict the value of $k$ to $3$. Similarly, we get the top-$3$ subframes for each news article using the guided LDA model. We randomly sampled $10$ articles from each side (left, right) for each topic resulting in $60$ articles, and identified their top-$3$ subframes using our model and guided LDA. We asked $2$ graduate students to annotate individually which set of subframes best describe the talking point of the news article. We found the Cohen's kappa score of $0.63$ which implies substantial agreement. In case of a tie, we break the tie by the 3rd annotator. While selecting the news articles, we considered news articles not having any of the seed indicators and having at least $300$ and at most $500$ words. Based on majority voting, In case of abortion, immigration and gun control respectively $16$, $18$ and $15$ news articles are better described using our model than guided LDA out of $20$ in each topic. In case of Immigration, $1$ news article had the same top $3$ subframes by both of the models.

\section{Analyzing Polarization on News Media} 
In this section we show how our model can be used to analyze polarization on news media. We focus on comparing several different qualitative results, contrasting the analysis obtained by policy frames and our subframe approach. Similar to the previous section, we identify the top 3 subframes in a news article using the embedding model. To compare with frame usage, we identify the top $3$ frames for each news article by following the same process used by \citet{field2018framing}, counting the number of word occurrences in an article from a frame lexicon and taking the most frequent $3$ as predicted frames.   


\begin{table}[t]
\begin{center}
 \scalebox{0.67}{
\begin{tabular}{>{\arraybackslash}m{2cm}|>{\centering\arraybackslash}m{3.5cm}|>{\centering\arraybackslash}m{4.2cm}}
 \hline
 \textsc{Topics} & \textsc{Frame Rank Corr.} & \textsc{Subframe Rank Corr.}\\
 \hline
 Abortion       & 0.86 (0.07)   & 0.25 (0.2)\\
 Immigration    & 0.81 (0.09)   & 0.54 (0.2)\\
 Gun Control    & 0.87 (0.08)   & 0.55 (0.3)\\
 \hline
\end{tabular}}
\end{center}
\begin{center}
 \scalebox{0.60}{\begin{tabular}{>{\arraybackslash}m{2cm}||>{\centering\arraybackslash}m{2cm}|>{\centering\arraybackslash}m{2cm}||>{\centering\arraybackslash}m{2cm}|>{\centering\arraybackslash}m{2cm}} 
 \hline 
\textsc{Topics} & \multicolumn{2}{c}{\textsc{\textbf{Left}}} & \multicolumn{2}{c}{\textsc{\textbf{Right}}}\\ [0.5ex]
\cline{2-5}
&   \textsc{Frame Rank Corr.}   &   \textsc{Subframe Rank Corr.}   &   \textsc{Frame Rank Corr.}   &   \textsc{Subframe Rank Corr.}\\
 \hline
 Abortion       & 0.92 (0.02) & 0.62 (0.2)  & 0.92 (0.05)   & 0.69 (0.08)\\
 Immigration    & 0.90 (0.05) & 0.78 (0.1)  & 0.85 (0.08)   & 0.61 (0.2)\\
 Gun Control    & 0.92 (0.03) & 0.76 (0.06) & 0.90 (0.04)   & 0.73 (0.15)\\
 \hline
 \end{tabular}}
\caption{Average Frame and Subframe rank correlation \textit{between} (top table) and \textit{within} (bottom table) ideologies calculated over years $2014$-$2019$. Standard deviations are in the brackets.}
\label{tab:frame-sf-rank-corr}
\end{center}
\end{table}

\begin{figure}[t!]
    \centering
    \begin{subfigure}[t]{0.23\textwidth}
        \centering
        \includegraphics[width=\textwidth]{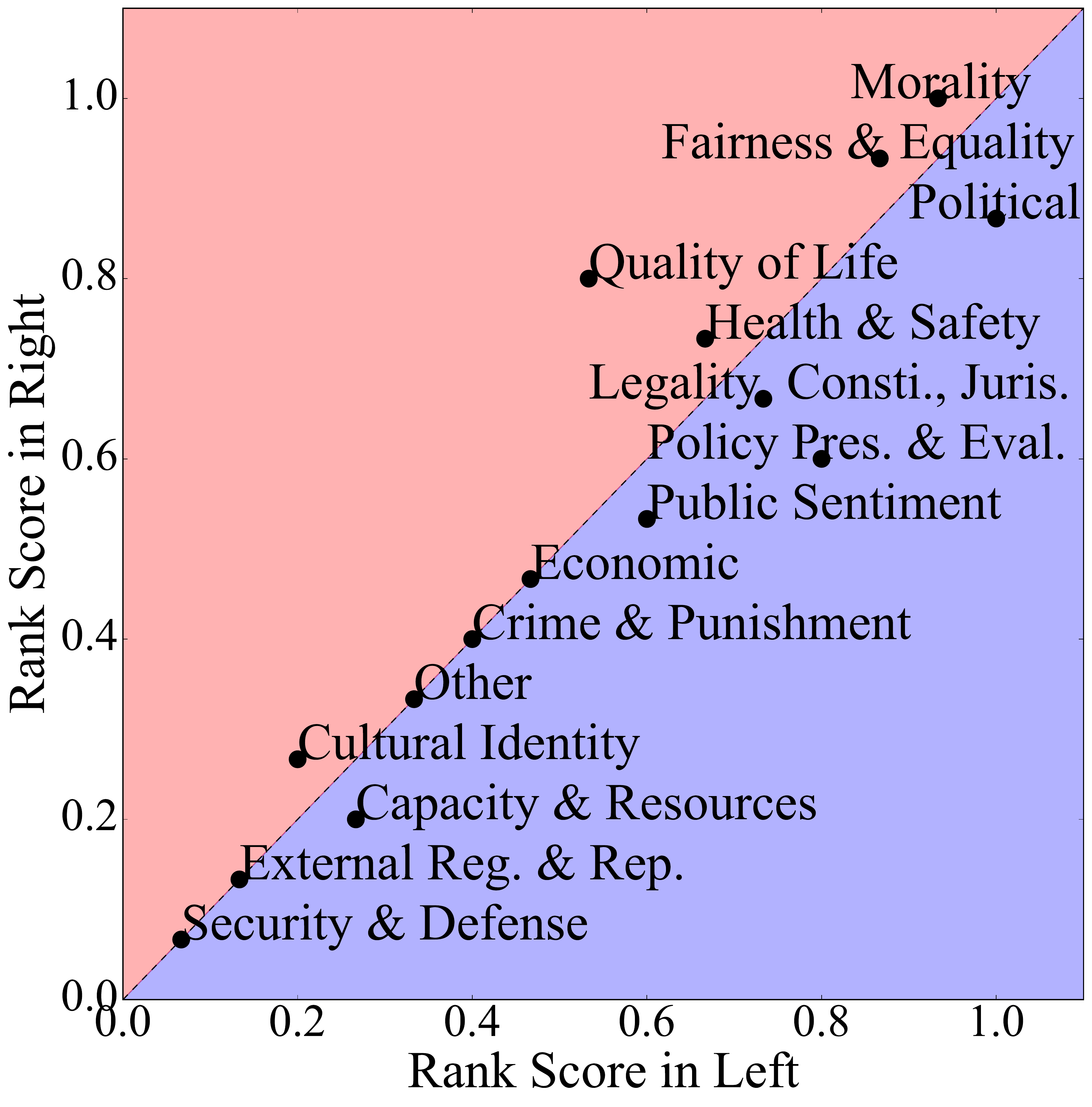}
        \caption{Frame Usage}
    \end{subfigure}%
    ~
    \begin{subfigure}[t]{0.23\textwidth}
        \centering
        \includegraphics[width=\textwidth]{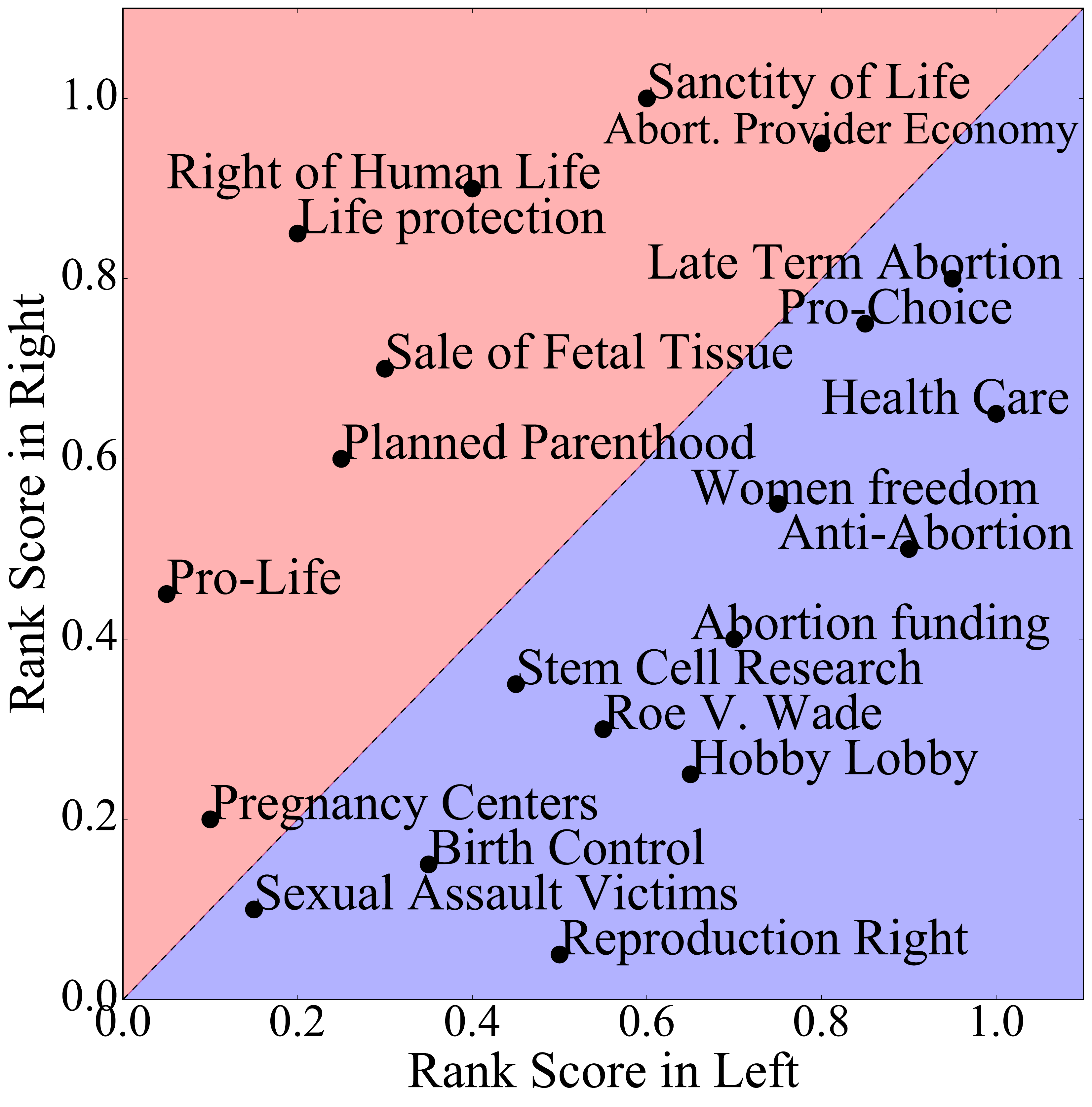}
        \caption{Subframe Usage}
    \end{subfigure}
    \caption{Polarization in usage of frames and subframes. The ranking scores are obtained by taking the normalized rank of the frames and subframes where the highest ranked instance get a score of 1. The rankings are over all news articles in the topic Abortion.}
    \label{fig:polarization}
\end{figure}

\begin{figure*}[t]
    \centering
    \begin{subfigure}[t]{0.4\textwidth}
        \centering
        \includegraphics[width=\textwidth]{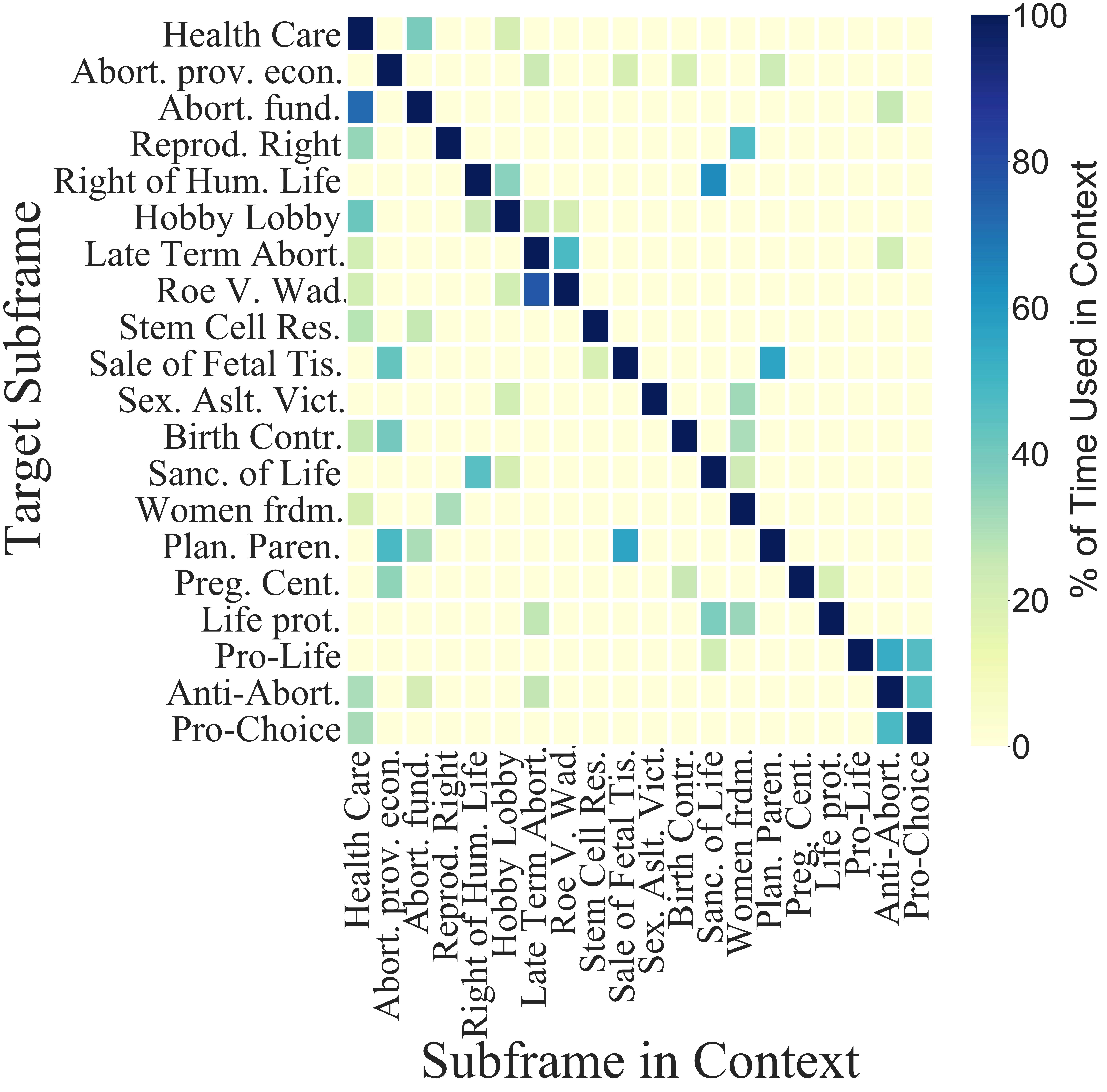}
        \caption{Left Articles}
    \end{subfigure}%
    ~ 
    \begin{subfigure}[t]{0.4\textwidth}
        \centering
        \includegraphics[width=\textwidth]{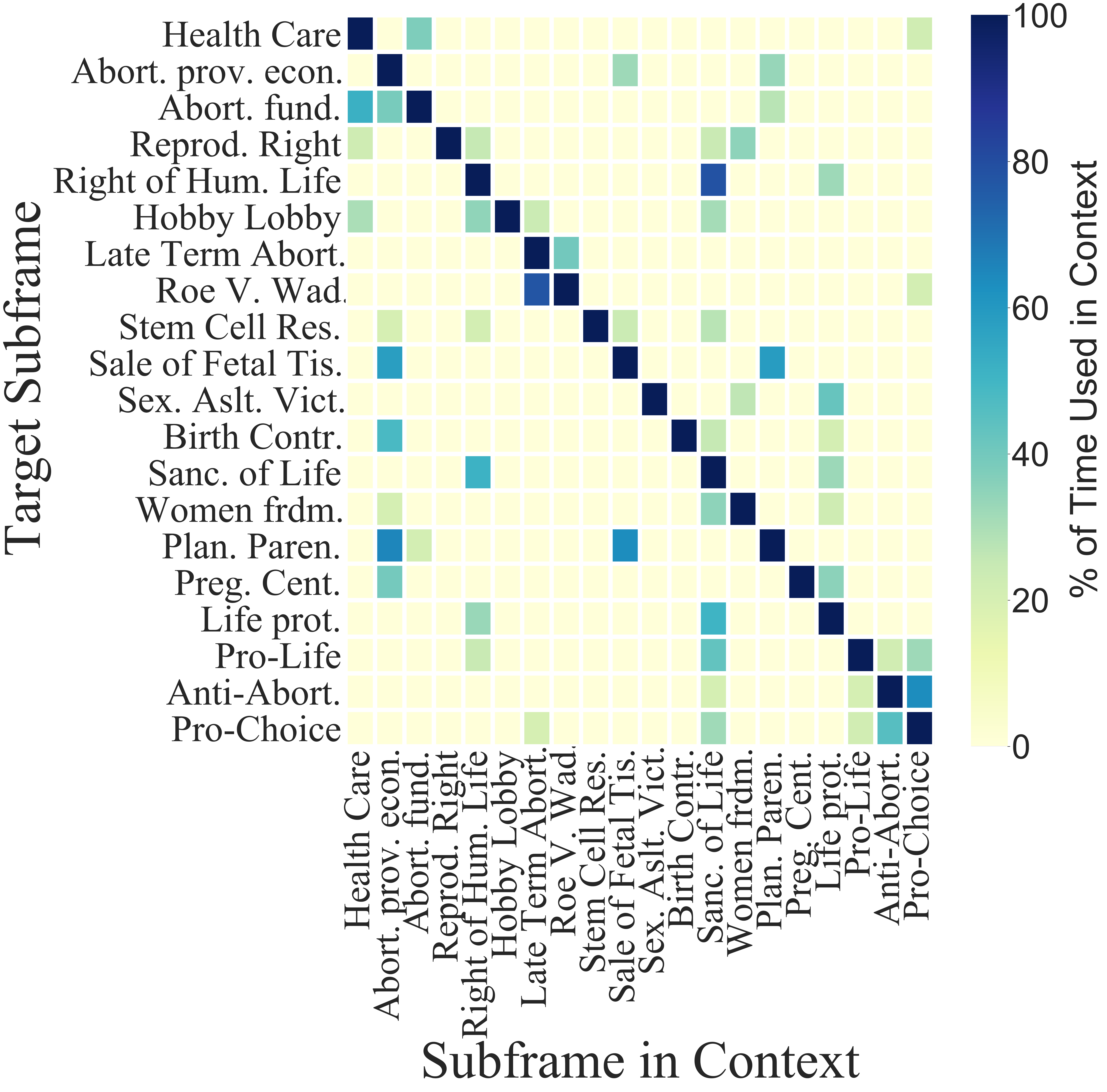}
        \caption{Right Articles}
    \end{subfigure}
    \caption{Heatmaps showing subframes used in the context of a subframe on the topic Abortion. Subframes used less than $20\%$ of the time in context are rounded down to zero for a cleaner representation purpose.}
    \label{fig:heatmaps}
\end{figure*}

\subsection{Overall Frame and Subframe Usage}
\label{subsection:between-within-corr}
To compare frame and subframe usage \textit{between} and \textit{within} ideologies, we create a ranked list of frames and subframes based on their occurrence frequency in articles identified with each ideology. We create ranked list for each year, and calculate the average correlation between the lists each year, \textit{between} ideologies to capture how similar are the framing dimensions, and \textit{within} ideologies by calculating the correlation between pairs of consecutive years within the same ideology, measuring the change in perspectives over time in each ideological camp. We use Spearman's Rank Correlation Coefficient to measure the agreement between rankings. Table \ref{tab:frame-sf-rank-corr} show the correlations averaged over years $2014$-$2019$. We take this time frame as it accounts for the majority of news articles in all $3$ topics (Table \ref{tab:dataset-summary}). Less agreement \textit{between} ideologies in subframe usage than frame usage shows that subframe analysis can better capture the polarization. Figure \ref{fig:polarization} shows this polarity for the topic \textit{abortion}. Subframes related to the fetus life is more used by the right while reproduction rights related subframes (Roe V. Wade, Women Freedom) are more focused by the left. In frame usage the parties are almost identical. This figure also shows that frames like `Security and Defense' are least used by each ideology which supports our claim in the subframe creation step that some frames are irrelevant to certain topics. Polarization graph for other two topics are shown in Appendix G. Less agreement in subframe usage than frame usage \textit{within parties} over years (in Table \ref{tab:frame-sf-rank-corr}) implies that parties use different subframes at different times, possibly in response to events occurred at that time. This hypothesis is further qualitatively analyzed in Section \ref{subsection:qualitative-analysis}.


\subsection{Subframes Instantiation Differences} 
To help get a better understanding of how subframes are used by each side, we analyze their co-occurrence. We represent this information using heatmaps. 
Each row in the heatmap captures the association strength between a given subframe (y-axis) and all the other subframes. The heatmap cell colors represents the percentage of times the two subframes appear in the same context.

The heatmaps for \textit{abortion} are shown in Figure \ref{fig:heatmaps} for both of the sides, and demonstrates how subframes are used differently by each side.  For example, when the left talks about `Women Freedom', it is used in the context of `Reproduction Right' and `Health Care',  implying the association between concepts. On the other hand, the right uses `Sanctity of Life', `Life Protection' in the context of `Women Freedom', implying they counter the issue of women freedom with the necessity of protecting lives of the unborn babies. In case of discussing `Hobby Lobby', the left relates it with `Roe V. Wade', possibly their conflicts, while the right relates `Hobby Lobby' with life protection issues more. 
Heatmaps for other two topics are shown in Appendix F.


\begin{table*}[t!]
\begin{center}
 \scalebox{0.52}{\begin{tabular}{>{\arraybackslash}m{3.1cm}|>{\arraybackslash}m{3.2cm}|>{\arraybackslash}m{3.3cm}||>{\arraybackslash}m{3.3cm}|>{\arraybackslash}m{3.3cm}||>{\arraybackslash}m{5.7cm}|>{\arraybackslash}m{5.7cm}} 
 \hline 
 
\textsc{Events} & \multicolumn{2}{c}{\textsc{\textbf{Frame Usage}}} & \multicolumn{2}{c}{\textsc{\textbf{Subframe Usage}}} & \multicolumn{2}{c}{\textsc{\textbf{Words Usage in Context of Common Subframes}}}\\ [0.5ex]
\hline
 &   \textsc{Left}   &   \textsc{Right}   &   \textsc{Left}   &   \textsc{Right} & \textsc{Left} & \textsc{Right}\\
\cline{2-7}
    \makecell[l]{\textbf{Abortion Event:}\\Leaked video of\\Planned Parenthood\\(Jul 14, 2015),\\Shooting at\\Planned Parenthood,\\CO (Nov 27, 2015)}& 
    \makecell[l]{- Political\\- Fairness \& Equality\\- Health \& Safety\\- Morality\\- Legal., Cons., Juri.}& 
    \makecell[l]{- Quality of Life\\- Morality\\- Fairness \& Equality\\- Health \& Safety\\- Political}& 
    \makecell[l]{- Planned Parenthood\\- Sale of Fetal Tissue\\- Abort. prov. econ.\\- Abortion funding\\- Women freedom}& 
    \makecell[l]{- Sale of Fetal Tissue\\- Abort. prov. econ.\\- Planned Parenthood\\- Sanctity of Life\\- Right of Hum. Life}& 
    \makecell[l]{- Sale of Fetal Tissue: \textit{sting, donation,}\\\textit{deceptively, state, health}\\- Planned Parenthood: \textit{called,}\\\textit{shooting, smear, campaign, spring}\\- Abortion Providers economy: \textit{first,}\\\textit{patient, state, go, take}} &
    \makecell[l]{- Sale of Fetal Tissue: \textit{story, organ,}\\\textit{harvesting, human, money}\\- Planned Parenthood: \textit{report, made,}\\\textit{service, affiliate, year}\\- Abortion Providers economy: \textit{gover-}\\\textit{nment, industry, profit, affiliate, claim}}\\
    \hline
    \hline
    
    \makecell[l]{\textbf{Imm. Event:}\\Midterm Election\\(Nov 6, 2018)\\Govt. Shutdown\\ (Dec 22, 2018 -\\Jan 25, 2019)}& 
    \makecell[l]{- Political\\- Crime \& Punish.\\- Ext. Reg. \& Rep.\\- Capacity \& Resour.\\- Fairness \& Equality}& 
    \makecell[l]{- Crime \& Punish.\\- Security \& Defense\\- Capacity \& Resour.\\- Political\\- Ext. Reg. \& Rep.}& 
    \makecell[l]{- Racism \& \\Xenophobia\\- Border Protection\\- Racial Identity\\- Family Sep. Policy\\- Detention}& 
    \makecell[l]{- Refugee\\- Border Protection\\- Deportation: Illegal\\ Immigrants\\- Asylum\\- Detention}& 
    \makecell[l]{- Border Protection: \textit{work, crisis,}\\\textit{agent, also, part}\\- Detention: \textit{mother, administration,}\\\textit{separated, woman, report}} &
    \makecell[l]{- Border Protection: \textit{week, migrant,}\\\textit{congress, illegal, secure}\\- Detention: \textit{release, county, bed,}\\\textit{officer, migrant}}\\
    \hline
    \hline
   
    \makecell[l]{\textbf{Gun Event:}\\Stoneman Douglas\\High School shoo-\\ting (Feb 14, 2018)}& 
    \makecell[l]{- Political\\- Crime \& Punish.\\- Fairness \& Equality\\- Quality of Life\\- Policy Pres., Eval.}& 
    \makecell[l]{- Political\\- Crime \& Punish.\\- Policy Pres., Eval.\\- Fairness \& Equality\\- Quality of Life}& 
    \makecell[l]{- School Safety\\- Gun Show Loophole\\- Gun Control to\\Restrain Violence\\- White Identity\\- Stop Gun Crime}& 
    \makecell[l]{- School Safety\\- Gun Show Loophole\\- Gun Control to\\Restrain Violence\\- Mental Health\\- Background Check}& 
    \makecell[l]{- School Safety: \textit{officer, elementary,}\\\textit{arming, classroom, time}\\- Gun Con. to Rest. Vio.: \textit{shot,}\\\textit{style, health, petition, expansion}\\- Gun Show Loophole: \textit{universal,}\\\textit{minimum, anyway, still, style}} &
    \makecell[l]{- School Safety: \textit{massacre, president,}\\\textit{staff, rifle, person}\\- Gun Con. to Rest. Vio.: \textit{said,}\\\textit{empower, extreme, danger, stop}\\- Gun Show Loophole: \textit{south, limit,}\\\textit{allowed, student, close}}\\
    \hline
\end{tabular}}
\caption{In response to real life events usage of frames, subframes and words by ideologies; all appearing in order of their rank by frequency of usage. Articles on topic Abortion are taken from 6 months period from the planned parenthood video leaking; on topic Immigration from Jul 1, 2018 to Jan 31, 2018; on topic Gun Control 1 month period from the shooting date. In case of Abortion, we don't consider the subframes 'pro-life', 'anti-abortion' and 'pro-choice' while ranking as they capture addressing framing.}
\label{tab:event-response}
\end{center}
\end{table*}

\subsection{Differences in Event News Coverage}
\label{subsection:qualitative-analysis}
To investigate how event news coverage differs across ideological lines,  we pick $3$ defining events, one in each topic, and investigate the usage of frames and subframes by either side around the time of those events. The events are as follows\footnote{The Wikipedia links to the events are in Appendix D.}.

\begin{itemize}
    \item \textbf{Abortion Event:} Undercover videos released on July 14, 2015 showing an official at Planned Parenthood discussing how to abort a fetus and preserve the organs and the costs associated with sharing that tissue with scientists. These videos and the defunding of Planned Parenthood came in presidential candidates' debates. Following this event, on Nov 27, 2015, three people were murdered at a Planned Parenthood health center in Colorado by a shooter.
    \item \textbf{Immigration Event:} In the 2018 US midterm elections (Nov 6, 2018), $40$ seats flipped from Republican to Democratic control. The election had a huge anti-immigration rhetoric. Following the election the longest government shutdown in US history occurred (Dec 22, 2018 - Jan 25, 2019), caused by a dispute over the funding amount for an expansion of the US-Mexico border barrier.
    \item \textbf{Gun Control Event: }On February 14, 2018, a gunman opened fire with a semi-automatic rifle at Marjory Stoneman Douglas High School in Parkland, Florida, killing 17 people and injuring 17 others. The shooter was a former student of the same school who is ethnically white.
\end{itemize}

\begin{figure*}[t]
    \centering
    \begin{subfigure}[t]{0.85\textwidth}
        \centering
        \includegraphics[width=\textwidth]{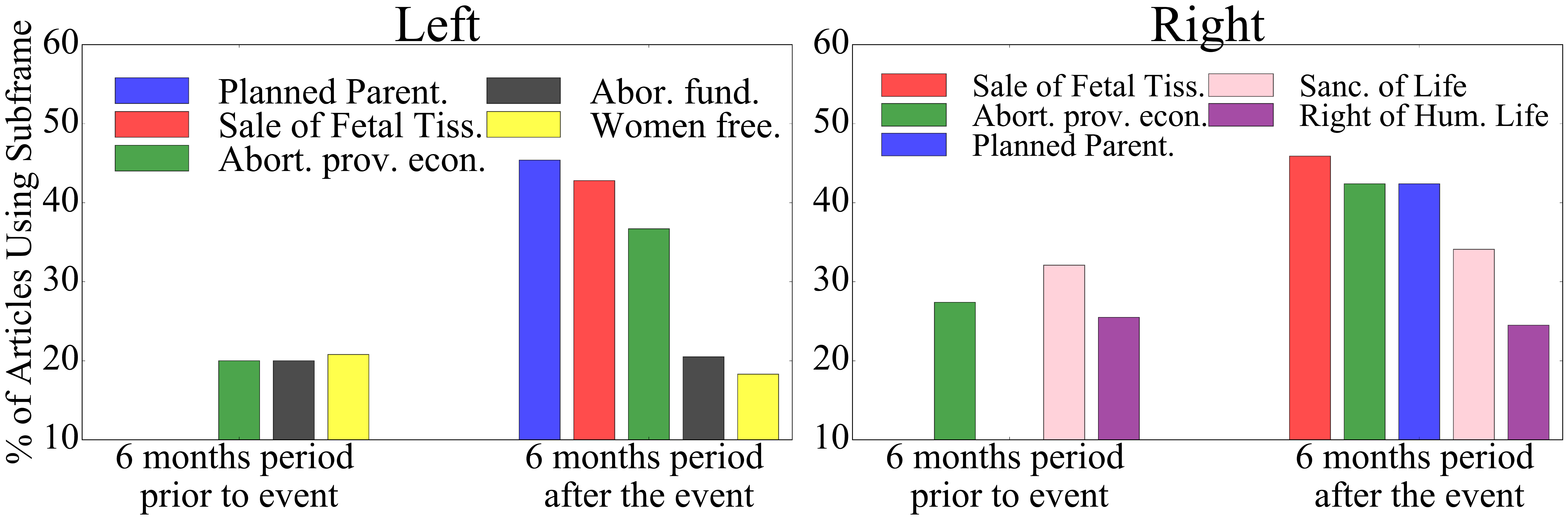}
        \caption{Abortion}
    \end{subfigure}%
    \\
    \begin{subfigure}[t]{0.85\textwidth}
        \centering
        \includegraphics[width=\textwidth]{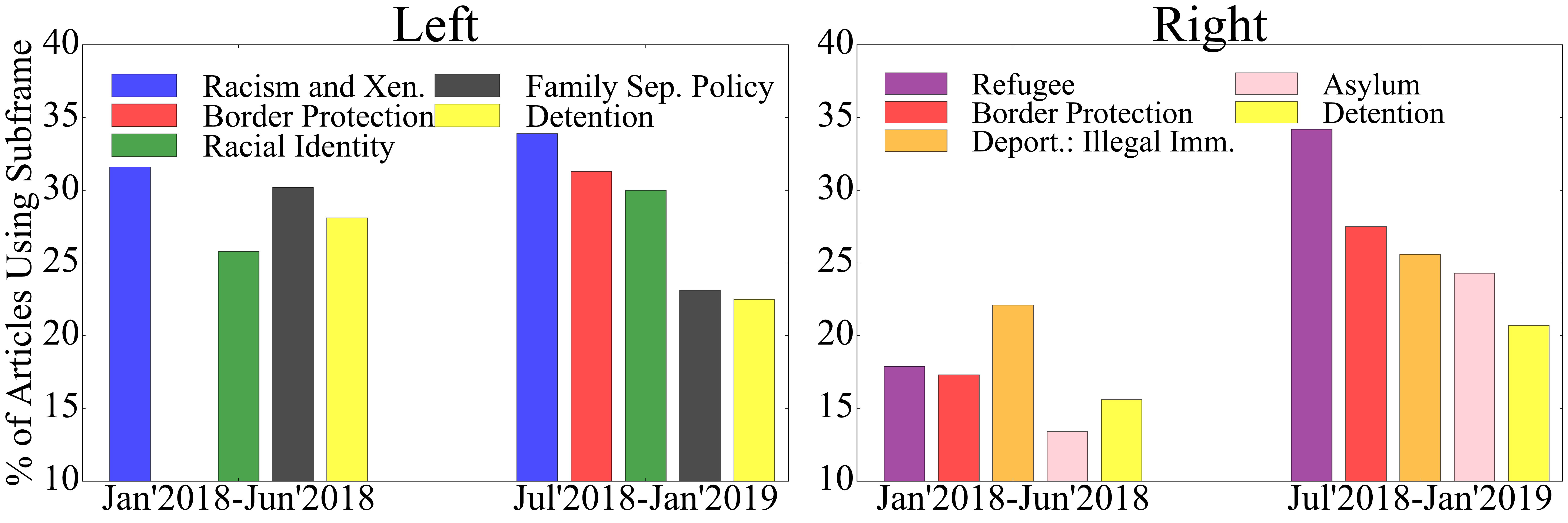}
        \caption{Immigration}
    \end{subfigure}%
    \\
    \begin{subfigure}[t]{0.85\textwidth}
        \centering
        \includegraphics[width=\textwidth]{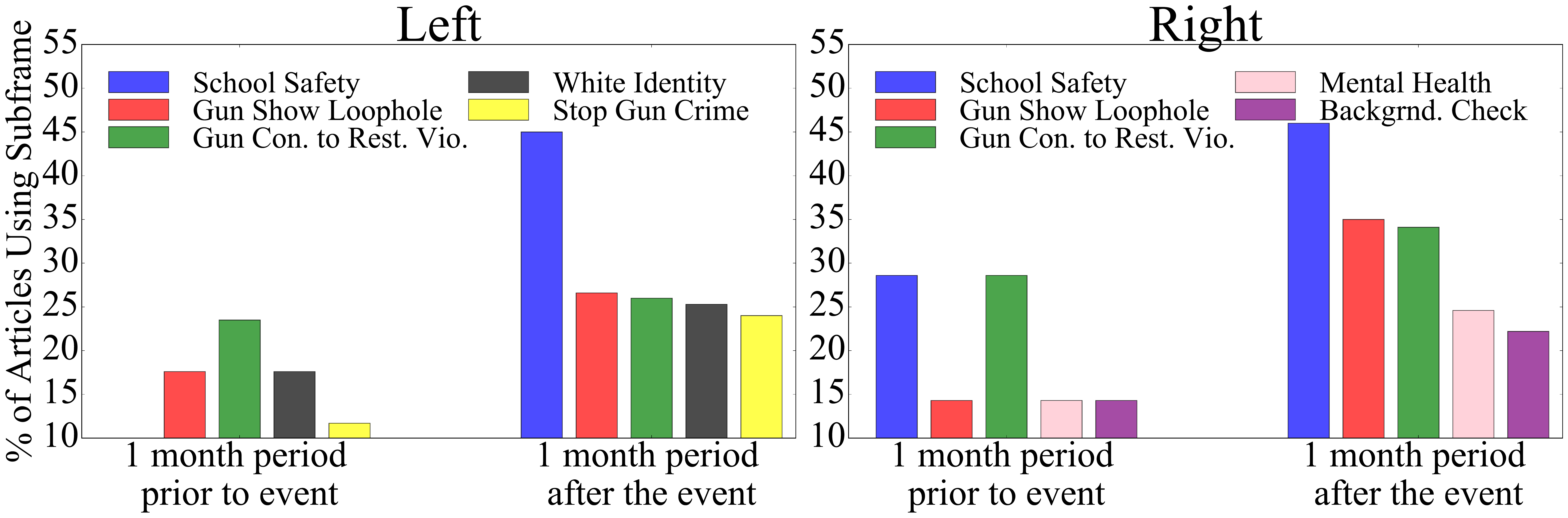}
        \caption{Gun Control}
    \end{subfigure}%
    \caption{Each party's use of the top-5 subframes before and after the events in Table \ref{tab:event-response}.}
    \label{fig:temporal-event}
\end{figure*}


The usage of frames and subframes in response to these events are summarized in Table \ref{tab:event-response}. It shows that frames usage by both parties overlaps, although with varying importance, and as a result offers limited insight. Analyzing subframes usage shows that some subframes are unique to each side in the context of the analyzed event. For example, the Abortion event analysis shows that `Sale of Fetal Tissue', `Planned Parenthood' and `Abortion Providers Economy' are used by both sides, as they are very relevant to the event. However, `Women freedom', and 'Sanctity of human life' are unique to each side. 

Figure \ref{fig:temporal-event} captures the change in framing behavior as a response to the event by comparing subframe usage before and after the event. It shows a spike in event related subframes by both sides, showing they respond to the specifics of the event. That supports our claim in Section \ref{subsection:between-within-corr} that the left and right react to events by using event related subframes. In topic Abortion, apart from the event related subframes, the left used `Abortion Funding' and `Women Freedom', and the right used subframes related to saving the life of the unborn. In the case of immigration, the left responded to the debates related to `Border Protection' by framing it as `Family Separation Policy', `Racism and Xenophobia', while the  right framed it as a `Refugee' issue. Around the school shooting event, the left talked about the `White Identity' of the shooter, while the right framed it as a `Mental Health' issue, a finding consistent with the claim by \citet{demszky2019analyzing} that the shooter's race plays a role in frame usage after a mass shooting event. 

We take a closer look at the difference between the usage of the same subframe by both sides by comparing the words used by each side. We look at the top $5$ high PMI words in the context of those subframes for each party\footnote{PMI calculation details is in Appendix E.} shown in the right-most two columns of Table \ref{tab:event-response}. Interestingly, in context of `Sale of Fetal Tissue', the left used words like \textit{sting, donation, deceptively}; suggesting they framed it as a propaganda ploy, while the right used words like \textit{organ, harvesting, human, money}; indicating a different interpretation. In case of `Border Protection', the left used words like \textit{crisis}, while the right used \textit{illegal, secure}. This analysis indicates again that even when the sides use similar event-specific subframes, their intent is different.

\section{Summary}\label{sec:summary}
 We study the news media coverage of $3$ politically polarized topics - \textit{abortion, immigration}, and \textit{gun control}; by breaking the high level policy frames into more fine grained, and topic-specific, subframes. We demonstrate that the subframes can account better for the way issues are framed in the news by both sides to influence their readers. We propose a novel embedding-based model extending our subframe lexicon to new text, allowing us to perform analysis more broadly. Our study serves as a starting point for additional work on hierarchical framing classification that can combine issue-specific or event-specific framing analysis with generalized framing dimensions that are comparable across different events and issues. To assist this effort, and improve reproducibility we provide additional details in Appendix H.

\section*{Acknowledgements}\label{sec:acknowledgements} We gratefully acknowledge Abida Sanjana Shemonti for helping in human evaluation and the anonymous reviewers for their insightful comments. 

\clearpage

\bibliography{refs}
\bibliographystyle{acl_natbib}
\clearpage
\appendix
\newpage

\section{Polarity in Frame Indicators and Subframe Indicators}
\label{Appendix-polarity-sf-f-indicators}
The rank correlation coefficient of frame indicator ranks and subframe indicator ranks between left and right for each issue frame are shown in Table \ref{tab:polarization-appendix}.

\begin{table*}[ht]
\begin{center}
 \scalebox{0.71}{\begin{tabular}{>{\arraybackslash}m{6.5cm}||>{\centering\arraybackslash}m{1.9cm}|>{\centering\arraybackslash}m{1.9cm}||>{\centering\arraybackslash}m{1.9cm}|>{\centering\arraybackslash}m{1.9cm}||>{\centering\arraybackslash}m{1.9cm}|>{\centering\arraybackslash}m{1.9cm}} 
 \hline 
\textsc{Frames} & \multicolumn{2}{c}{\textsc{\textbf{Abortion}}} & \multicolumn{2}{c}{\textsc{\textbf{Immigration}}} & \multicolumn{2}{c}{\textsc{\textbf{Gun Control}}}\\ [0.5ex]
\cline{2-7}
&   \textsc{Frame indicator}   &   \textsc{Subframe indicator}   &   \textsc{Frame indicator}   &   \textsc{Subframe indicator} &   \textsc{Frame indicator}   &   \textsc{Subframe indicator}\\
 \hline
 Economic                                   & 0.95&   0.44&   0.92& 0.28&   0.94&   0.29\\
 \hline
 Capacity and Resources                     & 0.95&   0.24&   0.89& 0.38&   0.92&   0.38\\
 \hline
 Morality                                   & 0.91&   0.30&   0.89& 0.16&   0.93&   0.16\\
 \hline
 Fairness and Equality                      & 0.93&   0.37&   0.92& 0.30&   0.93&   0.41\\
 \hline
 Legality, Constitutionality, Jurisdiction  & 0.96&   0.54&   0.92& 0.46&   0.95&   0.39\\
 \hline
 Policy Prescription and Evaluation         & 0.97&   0.58&   0.92& 0.07&   0.95&   0.48\\
 \hline
 Crime and Punishment                       & 0.94&   0.36&   0.90& 0.32&   0.94&   0.61\\
 \hline
 Security and Defense                       & 0.95&   0.17&   0.91& 0.28&   0.93&   0.41\\
 \hline
 Health and Safety                          & 0.96&   0.43&   0.94& 0.13&   0.95&   0.48\\
 \hline
 Quality of Life                           & 0.94&   0.34&   0.92& 0.37&   0.93&   0.19\\
 \hline
 Cultural Identity                         & 0.93&   0.22&   0.88& 0.03&   0.93&   0.41\\
 \hline
 Public Sentiment                          & 0.91&   0.33&   0.90& 0.05&   0.94&   0.42\\
 \hline
 Political                                 & 0.94&   0.53&   0.93& 0.51&   0.96&   0.46\\
 \hline
 External Regulation and Reputation        & 0.93&   0.22&   0.88& 0.18&   0.93&   0.51\\
 \hline
 Other                                     & 0.94&   0.18&   0.92& 0.21&   0.93&   0.38\\
 \hline
 Overall                                        & 0.94 (0.017)&   0.35 (0.128)&   0.91 (0.018)& 0.25 (0.142)&  0.94 (0.011)&   0.40 (0.112)\\
 \hline 
 \end{tabular}}
\caption{After ranking frames and sub-frame lexical indicators based on their usage in left and right biased documents, we measured and compared the correlation between the two ranks. Overall, frame indicators are not indicative of the label, as opposed to subframe indicators.}
\label{tab:polarization-appendix}
\end{center}
\end{table*}

\section{Subframe Seeds}
\label{Appendix:subframe-seeds}
The stemmed version of the following seeds were used for the corresponding subframes.

\subsection{Abortion}
\paragraph{Health Care} care act, affordable care act, affordable care
\paragraph{Abortion Provider Economy} abortion giant, abortion vendor, abortion business, abortion industry, largest abortion provider, sell abortion, human capital
\paragraph{Abortion Funding} fund abortion, abortion fund, funding of planned, fund family planning, title x funding, funding to planned, taxpayer funded abortion, parenthood's funding, family planning fund, funding of abortion, fund planned parenthood, fund for planned, funding for abortion, fund from planned, subsidizing abortion, cut planned parenthood, strip planned parenthood
\paragraph{Reproduction Right} reproductive justice, reproductive freedom, reproductive decision, reproductive choice, reproductive justice advocacy
\paragraph{Right of Human Life} life liberty, unborn life, life matter movement, respect for life, human life begin, right to life
\paragraph{Hobby Lobby} hobby lobby, hobby lobby case, freedom restoration, freedom restoration act, restoration act, exercise of religion
\paragraph{Late Term Abortion} partial birth, ban on partial, called partial birth, partial birth abortion
\paragraph{Roe V. Wade} revisit roe, landmark roe, decision roe, challenge roe, overturn roe, overrule roe, roe v, uphold roe, roe decision, see roe, decision in roe, court overturn roe, challenge to roe, ruling in roe, court's roe, roe is overturned, roe v wade, wade ruling, wade supreme, wade decision, v wade, wade supreme court, wade the landmark, wade the supreme
\paragraph{Stem Cell Research} cell research, tissue research, stem cell research, fetal tissue research
\paragraph{Sale of Fetal Tissue} sell fetal, sell baby, parenthood sell, sell fetal tissue, sell baby part, planned parenthood sell, procurement company, tissue procurement company
\paragraph{Sexual Assault Victims} rape victim, statutory rape, forced rape, victim of rape, sex crime, child sex, sex trafficking, sexual abuse, sexually offend, sexual assault, sexual misconduct, sex predator, child sex abuse, accused of sexual, victim of sexual, trafficking victim, sex trafficker', human trafficking
\paragraph{Birth Control} birth control, birth control pill, cover birth control, use birth control, birth control mandate, birth control access, unwanted pregnancy, prevent unwanted, prevent unwanted pregnancy, prevent unintended, prevent pregnancy, drug induced, abortion inducing drug, drug induced abortion
\paragraph{Sanctity of Life} sanctity of life, life is sacred, believe life, life catholicism, priest for life, evil of abortion, abortion is murdering, abortion is wrong
\paragraph{Women Freedom} punish woman, hurt woman, control woman, force woman, woman's movement
\paragraph{Planned Parenthood} parenthood support, planned parenthood clinic, planned parenthood abortion, planned parenthood sting, local planned parenthood, planned parenthood support
\paragraph{Pregnancy Centers} pregnancy help center, pregnancy resource center, resource center, pregnancy center, pregnancy help, pregnancy resource
\paragraph{Life Protection} child protection, protect life, baby's life, child's life, take a life, baby's life, end a life, take the life, pro life pregnancy, end the life, end of life, kill the baby, abort the baby, rip the baby, child killing, kill the child
\paragraph{Pro-life} pro life vote, pro life message, pro life group, pro life campaign, pro life advocate, pro life organization, pro life candidate, strong pro life, life rally, life protest, life voter, life group, life campaign, life organization, life candidate, life message, life advocate, life supporter, public life, life commission, coalition for life, march for life
\paragraph{Anti-abortion} anti abortion protest, anti abortion, anti abortion march, anti abortion right, anti abortion vote, anti abortion organization, anti abortion protest, anti abortion group, anti abortion democrat, anti abortion candidate, anti abortion advocate, anti abortion position, anti abortion lawmaker, opposed to abortion, oppose abortion right, oppose abortion, oppose abortion right, opponent of abortion
\paragraph{Pro-choice} pro choice vote, pro abortion right, pro choice organizing, pro choice position, pro choice woman, pro choice group, pro abortion group, pro choice candidate, pro choice advocate, supported abortion right, defend abortion right, support of abortion, favor abortion right, abortion right supporter

\subsection{Immigration}

\paragraph{Minimum Wage} income inequality, raise the minimum, minimum wage
\paragraph{Salary stagnation} cut salary, wage cut, stagnant wage, wage stagnation, lowering wage, wage lowering, driven down wage
\paragraph{Wealth Gap} widen wealth, shift wealth, wealth gap, wealth from young, widen wealth gap, immigration shift wealth, price widen wealth, wealth gap reduce
\paragraph{Cheap labor availability} cheap labor, cheap labor economy, low wage work, cheap foreign worker, cheap foreign labor, cheap labor migration, cheap foreigner, massive cheap labor, cheap labor strategy, successful cheap labor, cheap worker, inflow of cheap, cheap labor policy
\paragraph{Taxpayer Money} pay tax, taxpayer money, taxpayer dollar
\paragraph{Deportation: Illegal Immigrants} deport illegal, deport illegal immigrant, deport illegal alien, deportation of illegal, previously deported illegal, deport undocumented
\paragraph{Deportation: In General} face deportation, deport immigrant, deport back, deport person, mas deportation, deport million, deportation policy, arrest and deportation, detain and deport, stop the deportation
\paragraph{Detention} federal detention, immigrant detention, ice detention, detention facility, detention center, immigrant detention facility, release from detention, immigrant detention center
\paragraph{Terrorism} foreign terrorist, potential terrorist, terrorism related, terrorist suspect, terrorist problem, terrorist threat, suspected terrorist, terrorist group, terrorist organization, terrorist activity, domestic terrorism, war on terrorism
\paragraph{Border Protection} porous border, border fencing, border barrier, border enforcement, build the border, united state border, illegal border crossing, cross my border, border wall construction, wall prototype, build wall, build the wall, secure fencing, mile of fence
\paragraph{Asylum} grant asylum, asylum case, asylum application, asylum applicant, legitimate asylum, deny asylum, asylum claim, asylum rule, asylum officer, political asylum, asylum process, asylum seeking, seek asylum, asylum request, asylum system, asylum law, asylum hearing, claim asylum, asylum policy, qualified for asylum, claim for asylum, eligibility for asylum, apply for asylum, person seek asylum, refuge and asylum, number of asylum, immigration and asylum, ask for asylum
\paragraph{Refugee} refugee status, seek refugee, refugee and asylum
\paragraph{Birthright citizenship and 14th Amendment} birthright citizenship, end birthright, end birthright citizenship, fourteenth amendment, 14th amendment, automatically citizen
\paragraph{Amnesty} grant amnesty, executive amnesty, temporary amnesty, expand amnesty, amnesty program, given amnesty, amnesty bill, offer amnesty, amnesty proposal, amnesty plan, act amnesty, amnesty legislation, temporary amnesty program, amnesty to illegal
\paragraph{Dream Act} dreamer illegal, dream act, dream act amnesty
\paragraph{Family Separation Policy} separation policy, separate family, family separation policy, policy of separation, separation of child, end family separation, practice of separating, separation of family
\paragraph{DACA} deferred action, created deferred action, era deferred action, childhood arrival, action for childhood, illegally as child, country as child, childhood arrival program
\paragraph{Racism and Xenophobia} race bait, racial discrimination, racist attack, racist profiling, racism and xenophobia
\paragraph{Merit Based Immigration} merit based, based on merit, merit based system, merit based immigration
\paragraph{Human Right} human right, human right abuse, human right advocate, human right violation, civil right, civil disobedience, civil liberty, civil right activist, civil right movement
\paragraph{Racial Identity} white national, white supremacist, white supremacy, class white, white male, white race, white person, white identity, white woman, white man, white worker, new black, black community, younger black, black man, black woman, black voter, black person, first black, black president, black and brown, black and white, person of color, non white
\paragraph{Born Identity} born outside, foreign born, international migrant, eastern refugee, foreign student, foreign refugee, foreign born population, number of foreigner, family based chain, based chain migration, illegal alien population

\subsection{Gun Control}

\paragraph{Gun Buyback Program} buyback second, higher buyback, buyback rate, lower buyback, buyback program, gun buyback, buyback second firearm, lower buyback rate, higher buyback rate, gun buyback program
\paragraph{Gun Business Industry} firearm industry, gun business, gun company, gun market, firearm manufacturer, gun manufacturer, gun industry, firearm dealer, gun shop, gun dealer, licensed dealer, gun shop owner, gun store owner, licensed firearm dealer
\paragraph{School Safety} arming school, school security, school safety, student safety, protect student, arming teacher
\paragraph{White Identity} white guy, white male, white person, white supremacy
\paragraph{Person of Color Identity} black male, black neighborhood, black person, black man, person of color
\paragraph{Ban on Handgun} transfer ban, handgun transfer, handgun transfer ban, handgun ban, ban on handgun
\paragraph{Second Amendment} heller decision, second amendment protected, second amendment guarantee, second amendment right, 2nd amendment right, amendment right, second amendment right, amendment protected, protection the second
\paragraph{Concealed Carry Reciprocity Act} reciprocity act, carry reciprocity, carry reciprocity act, concealed carry, carry concealed firearm, conceal carry law, concealed carry permit, concealed carry license
\paragraph{Gun Control to Restrain Violence} violence restraining, gun violence restraining, violence restraining order, domestic violence restraining, prevent gun violence
\paragraph{Illegal Gun} illegal possession, gun illegally, illegal gun, illegal firearm, criminal possession
\paragraph{Gun Show Loophole} loophole that allow, show loophole, close loophole, gun show
\paragraph{Background Check} instant criminal background, criminal background check, background check system, gun background check, strengthen background check, passing background check, stronger background check, new background check, conduct background check, background check requirement, strengthening background check, universal background check, comprehensive background check
\paragraph{Terrorist Attack} terrorist threat, international terrorism, terrorist watch, terrorist attack, terrorist suspect, terrorist group, terrorist activity, suspected terrorist, domestic terror, foreign terror, terrorist organization, anti terrorism, terror gap, terrorist watch list, war on terrorism, act of terrorism
\paragraph{Gun Research} gun violence research, gun death researcher, gun research, death researcher, violence researcher
\paragraph{Mental Health} seriously mentally, severe mental, mental state, mental illness, address mental, mental health care, person with mental, mentally ill person
\paragraph{Gun Homicide} firearm death, gun death, shooting death, gun death domestic, gun death, gun death rate, gun death researcher, reduce gun death, gun public health
\paragraph{Assault Weapon} new assault, ban assault, semiautomatic assault, new assault weapon, ban assault weapon, automatic firearm, automatic gun, fully automatic, automatic rifle, semiautomatic rifle, semi automatic, automatic fire, automatic machine, semiautomatic assault, semiautomatic weapon, allow semi automatic, fully automatic rifle, ban semi automatic, fully automatic firearm, full automatic weapon, semi automatic gun, semi automatic rifle, fully automatic machine, rifle and shotgun, rifle to fire, rifle ban
\paragraph{Right to Self-Defense} religious right, given right, god given right, right to protect, right of gun, god given, exercised their second, bill of right
\paragraph{Stop Gun Crime} commit violence, mas violence, history of violent, culture of violence, stop gun violence, risk of violence, curb gun violence, victim of violence, end gun violence, thought and prayer, victim of domestic

\section{Subframe Description}
\label{Appendix:subframe-description}
Subframes with corresponding description are summarized in Table \ref{tab:subframe-description-appendix}
\begin{table*}[ht]
\begin{center}
 \scalebox{0.67}{\begin{tabular}{>{\arraybackslash}m{1.8cm}|>{\arraybackslash}m{5cm}|>{\arraybackslash}m{15cm}} 
 \hline 
\multicolumn{3}{c}{\textsc{\textbf{Abortion}}}\\ [0.5ex]
\cline{1-3}
  \textbf{\textsc{Frame}} &    \textbf{\textsc{Subframe}}   &   \textbf{\textsc{Short Description}}\\
 \hline
 Economic   & Health Care & Affordable Care Act, healthcare facilities, health insurance, their coverage etc.\\
            \cline{2-3}
            & Abort. Provider Economy & Statistics, services, profits of abortion providers like Planned Parenthood.\\
            \cline{2-3}
            & Abortion Funding & Source of funding; granting or cutting funding for abortion providers like Planned Parenthood. \\
 \hline
 Fairness \&    & Reproduction Right & Reproduction rights and women's access to reproduction healthcare. \\
                \cline{2-3}
  Equality      & Right of Human Life & Fetus in the womb has the same right of life as a grown human.\\
 \hline
 Legality,          & Hobby Lobby & Supreme Court's exemption for corporations to provide contraceptives if it conflicts with their religious belief.\\
                    \cline{2-3}
 Consti.,           & Late Term Abortion & Discuss ban and regulation on abortion after later stages of pregnancy.\\
                    \cline{2-3}
 Jurisdiction       & Roe V. Wade & Implications of the 1973 landmark decision of the U.S. Supreme Court that ensures the right to choose.\\
 \hline
 Crime \&   & Stem Cell Research & Research and its implications using stem cell, embryonic cell and fetal tissue.\\
            \cline{2-3}
 Punishment & Sale of Fetal Tissue & Abortion providers donation or selling of the fetal tissue and baby body parts from aborted babies.\\
            \cline{2-3}
            & Sexual Assault Victims & Any kind of sexual offense against women and pregnancies resulted from that.\\
 \hline 
 Health, Saf.  & Birth Control & Birth control measures and access to those.\\
 \hline
 Morality   & Sanctity of Life & The holiness of life from a religious and moral perspective and the evil of abortion.\\ 
            \cline{2-3}
            & Women Freedom & Advocating women freedom or talking about suppression on women, from a moral perspective.\\
 \hline
 Quality            & Planned Parenthood & Abortion services provided by Planned Parenthood.\\
                    \cline{2-3}
 of Life            & Pregnancy Centers & Pregnancy services provided by pregnancy care centers, pregnancy crisis center etc.\\
                    \cline{2-3}
                    & Life Protection & Abortion kills human being and they should be protected.\\
 \hline
 Public             & Pro-life & Addressing of any personality, movement or legislation as supporting life.\\
                    \cline{2-3}
 Sentiment          & Anti-abortion & Addressing of any personality, movement or legislation as opposing abortion instead of addressing as pro-life.\\
                    \cline{2-3}
                    & Pro-choice & Addressing of any personality, movement or legislation as supporting abortion and the right to choose.\\
 \hline 
 \end{tabular}}

 \scalebox{0.67}{\begin{tabular}{>{\arraybackslash}m{1.8cm}|>{\arraybackslash}m{5cm}|>{\arraybackslash}m{15cm}} 
 \hline 
\multicolumn{3}{c}{\textsc{\textbf{Immigration}}}\\ [0.5ex]
\cline{1-3}
  \textbf{\textsc{Frame}} &    \textbf{\textsc{Subframe}}   &   \textbf{\textsc{Short Description}}\\
 \hline
 Economic   & Minimum Wage & Wage inequality and discussion on raising the minimum wage.\\
            \cline{2-3}
            & Salary Stagnation & Reasons of salary stagnation and how to overcome those. \\
            \cline{2-3}
            & Wealth Gap & Wealth gap among the classes in the society; profits by large organizations etc.\\
            \cline{2-3}
            & Cheap Labor Availability & Cheap labor availability and its effects.\\
            \cline{2-3}
            & Taxpayer Money & Taxpayer money and the facilities they get or are deprived of, such as social security.\\
 \hline
 Fairness \&    & Racism and Xenophobia & Addressing of someone/something racist and xenophobic in a discussion.\\
                \cline{2-3}
  Equality      & Merit Based Immigration & Discussion on merit based immigration system.\\
                \cline{2-3}
                & Human Right & Necessity of protecting human and civil rights; their violations.\\
 \hline
 Legality,          & Asylum & Implications of granting asylum to the asylum seeking migrants.\\
                    \cline{2-3}
 Consti.,           & Refugee & Political refugees from various countries.\\
                    \cline{2-3}
 Jurisdiction       & Birth citizenship \& 14th Amen. & Birthright citizenship; 14th Amendment; citizenship granting procedure.\\
 \hline
 Crime \&   & Deportation: Illegal Immigrants & Necessity of deportation of the illegal immigrants.\\
            \cline{2-3}
 Punishment & Deportation: In General & Procedure, policy and way to deport the undocumented immigrants.\\
            \cline{2-3}
            & Detention & Detention facilities; detention procedure and the state of the detainees.\\ 
 \hline 
 Security   & Terrorism & Threats of terrorism by foreign nationals.\\
            \cline{2-3}
 \& defense & Border Protection & Border wall; border patrol and other measures to secure the border.\\
 \hline
 Policy                 & Amnesty & Implications and procedure of granting amnesty to the undocumented immigrants.\\
                        \cline{2-3}
 Prescription,          & DREAM Act & 2001, DREAM Act, its implications; DREAMers and procedure of their path to citizenship.\\
                        \cline{2-3}
 Evaluation             & Family Separation Policy & Family separation policy and its effects; separation of children from their families in the border.\\
                        \cline{2-3}
                        & DACA & DACA policy that protects the individuals from deportation who came to the USA as children.\\
 \hline
 Cultural           & Racial Identity & Discussion on a topic by focusing on the race.\\
                    \cline{2-3}
 Identity           & Born Identity & Discussion on a topic by addressing the born identity, such as, `foreign born'.\\
 \hline 
 \end{tabular}}

 \scalebox{0.67}{\begin{tabular}{>{\arraybackslash}m{1.8cm}|>{\arraybackslash}m{5cm}|>{\arraybackslash}m{15cm}} 
 \hline 
\multicolumn{3}{c}{\textsc{\textbf{Gun Control}}}\\ [0.5ex]
\cline{1-3}
  \textbf{\textsc{Frame}} &    \textbf{\textsc{Subframe}}   &   \textbf{\textsc{Short Description}}\\
 \hline
 Economic   & Gun Buyback Program & Gun buyback program and its effects.\\
            \cline{2-3}
            & Gun Business & Licensed gun store owners; gun business industry.\\
 \hline
 Health \&  & Gun Research & Research on gun violence and how to control it; funding on gun research.\\
            \cline{2-3}
 Safety     & Mental Health & Mental illness; importance of providing mental health care.\\
            \cline{2-3}
            & Gun Homicide & Statistics on deaths due to gun violence.\\
 \hline
 Legality,          & Ban on Handgun & Banning handgun and its effects.\\
                    \cline{2-3}
 Consti.,           & Second Amendment & 2nd Amendment which ensures right to self-defense and allows law abiding citizens to carry guns.\\
                    \cline{2-3}
 Jurisdiction       & Concealed Carry Reciprocity Act & Concealed carry reciprocity act and its effects and implications.\\
                    \cline{2-3}
                    & Gun Control to Restrain Violence & Violence-restraining gun control measures.\\
 \hline
 Crime \&   & Illegal Gun & Illegal possession of gun; gun trafficking etc.\\
            \cline{2-3}
 Punishment & Gun Show Loophole & Loophole in the gun shows that allows criminals to get gun.\\
 \hline 
 Security   & Background Check & Necessity of background check and ways to ensure it while selling guns.\\
            \cline{2-3}
 \& defense & Terrorist Attack & Threats of terrorist attack.\\
 \hline
 Policy Pres., Eval.    & Assault Weapon & Debate over the definition of assault weapon and which ones are needed to be banned.\\
 \hline
 Cultural           & White Identity & Focusing on white racial identity of a person; white supremacy etc.\\
                    \cline{2-3}
 Identity           & Person of Color Identity & Focusing on person of color racial identity.\\
 \hline 
 Capacity, Resource & School Safety & Measures to ensure school safety; arming teachers; control gun to reduce violence in schools etc.\\
 \hline
 Morality   & Right to Self-Defense & God given right to self defense; necessity of carrying guns for self-defense etc.\\
            \cline{2-3}
            & Stop Gun Crime & Urge to stop gun violence; expression of solidarity with mass shooting victims etc.\\
 \hline
 \end{tabular}}
 
\caption{Subframe Description}
\label{tab:subframe-description-appendix}
\end{center}
\end{table*}

\section{Event References}
\label{Appendix:event-reference}
\begin{itemize}
    \item \textbf{Topic: Abortion}\\
    \textbf{Event: }
    Undercover videos released on July 14, 2015 showing an official at Planned Parenthood discussing how to abort a fetus and preserve the organs and the costs associated with sharing that tissue with scientists\footnote{\url{https://en.wikipedia.org/wiki/Planned\_Parenthood\_2015\_undercover\_videos\_controversy}}. These videos and defunding of Planned Parenthood came in presidential candidates' debates. Following this event, on Nov 27, 2015, three people were murdered at a Planned Parenthood health center in Colorado by a shooter\footnote{\url{https://en.wikipedia.org/wiki/Colorado\_Springs\_Planned\_Parenthood\_shooting}}.
    
    \item \textbf{Topic: Immigration:}\\
    \textbf{Event: }
    In the 2018 US midterm elections (Nov 6, 2018), $40$ seats flipped from Republican to Democratic control\footnote{\url{https://en.wikipedia.org/wiki/2018\_United\_States\_elections}}. The election had a huge anti-immigration rhetoric. Following the election the longest government shutdown in US history occurred (Dec 22, 2018 - Jan 25, 2019)\footnote{\url{https://en.wikipedia.org/wiki/2018\%E2\%80\%9319\_United\_States\_federal\_government\_shutdown}}, caused by a dispute over the funding amount for an expansion of the US-Mexico border barrier.
    
    \item \textbf{Topic: Gun Control:}\\
    \textbf{Event: }On February 14, 2018, a gunman opened fire with a semi-automatic rifle at Marjory Stoneman Douglas High School in Parkland, Florida, killing 17 people and injuring 17 others. The shooter was a former student of the same school who is ethnically white\footnote{\url{https://en.wikipedia.org/wiki/Stoneman\_Douglas\_High_School\_shooting}}.
\end{itemize}

\section{Detection of Top-$5$ Highest PMI Words in the Context of Subframes by for Each Party}
\label{Appendix:pmi-calculation}
To detect the top PMI words for each party label (left, right) in the context of a subframe, $s$, we follow the following procedure. If a news article is detected to have a subframe, $s$, in its top $3$ subframes, we take only the paragraphs from that news article which have the highest probability as having the subframe, $s$, over all the subframes and also belongs to the top-$500$ nearest paragraphs of the subframe, $s$, in the embeddings space. This top-$500$ list is created based on cosine similarity between the embedding of the subframe, $s$, and the embedding of all paragraphs on the topic. Now we tokenize the subset of paragraphs having subframe, $s$. To remove stopwords and very rarely occurring words we consider only the words appearing in less than $5\%$ of the paragraphs and more than $60\%$ of the paragraphs. For a word $w$ we calculate the pointwise mutual information (PMI) with label $l$, $I(w, l)$ using the following formula.
\begin{align*}
     I(w,l)=\operatorname{log}\frac{P(w|l)}{P(w)}
\end{align*}{}
Where $P(w|l)$ is computed by taking all paragraphs with label $l$ and computing $\frac{count(w)}{count(allwords)}$ and similarly, $P(w)$ is computed by counting word $w$ over the set of paragraphs combining both left and right biased ones. Now, we rank words for each label (left, right) based on their PMI scores.

\section{Usage of Subframes in Context of Other Subframes}
\label{Appendix:heatmaps}
\begin{figure*}[ht]
  \begin{subfigure}{\linewidth}
  \centering
  \includegraphics[width=.48\linewidth]{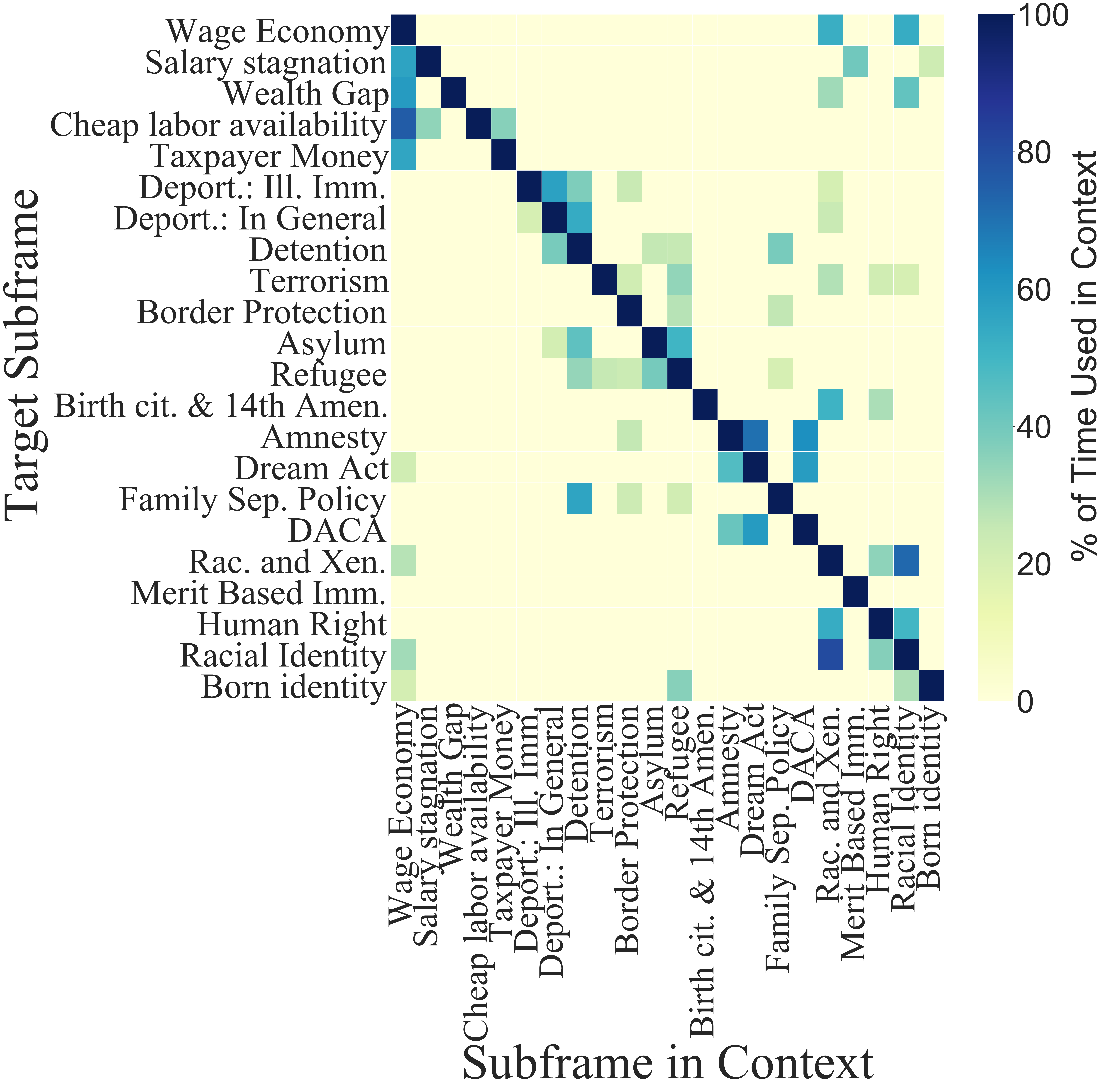}
  \includegraphics[width=.48\linewidth]{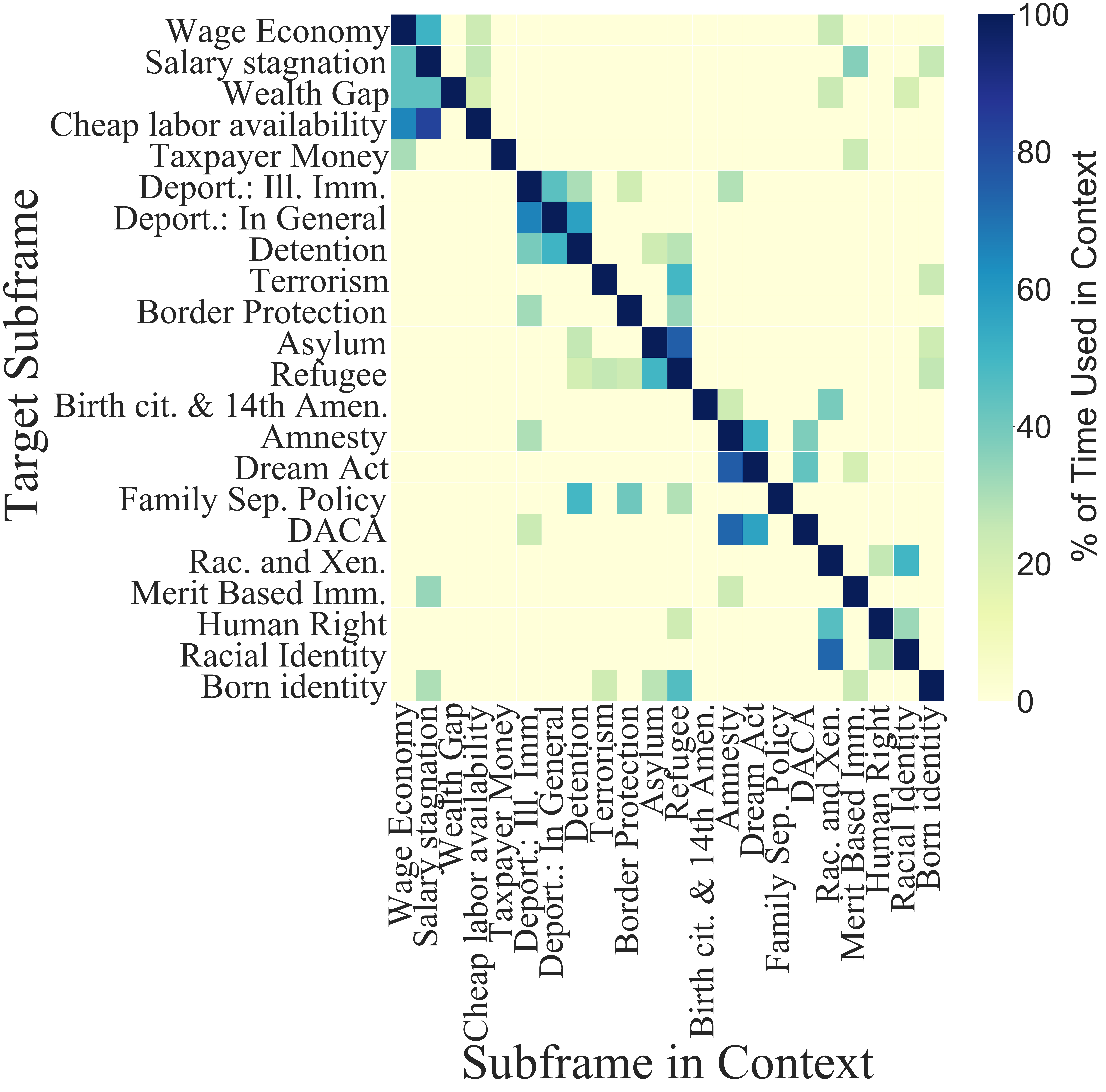}\\
  \caption{Immigration}
  \end{subfigure}\par\medskip
  \begin{subfigure}{\linewidth}
  \centering
  \includegraphics[width=.48\linewidth]{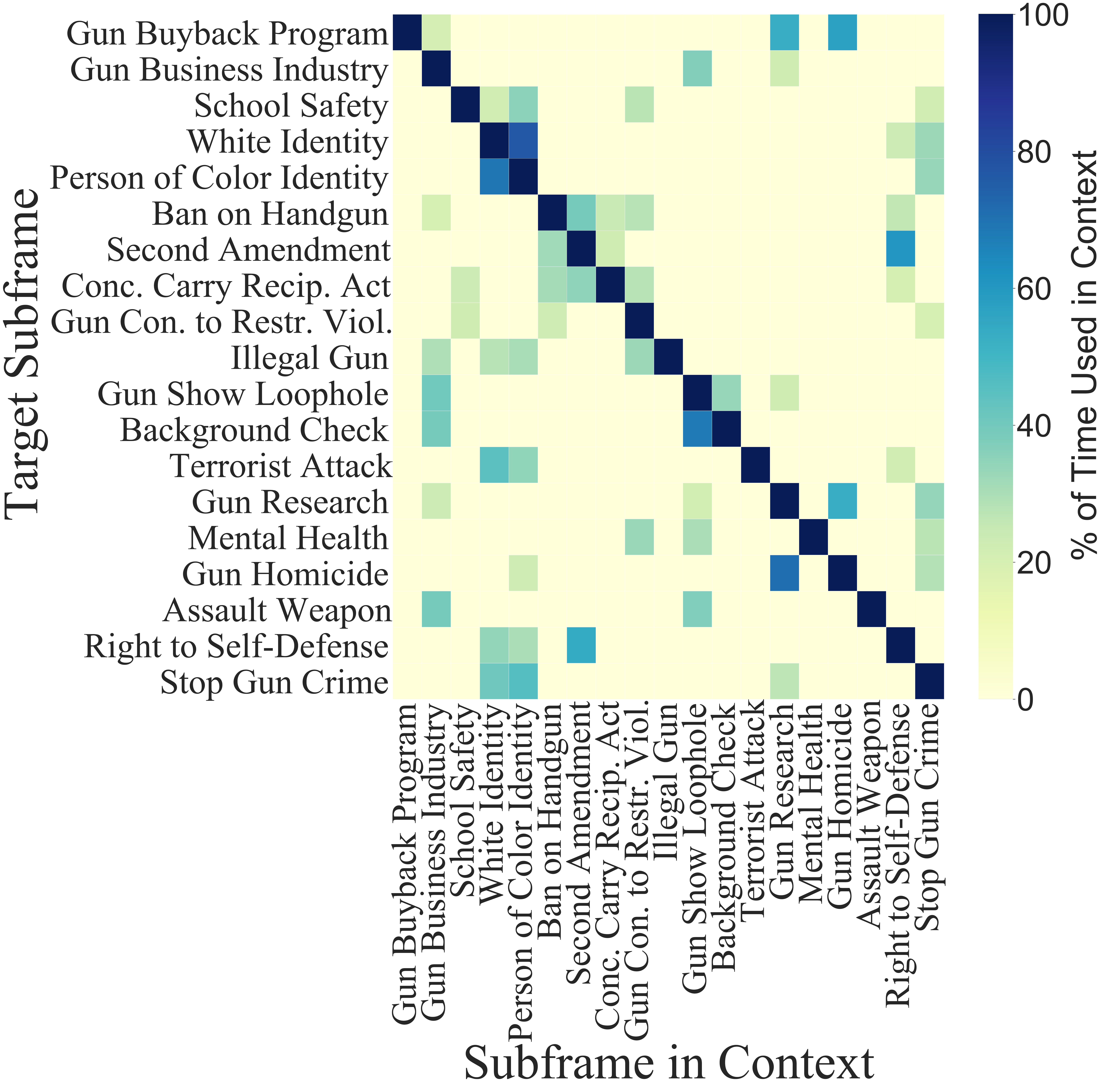}
  \includegraphics[width=.48\linewidth]{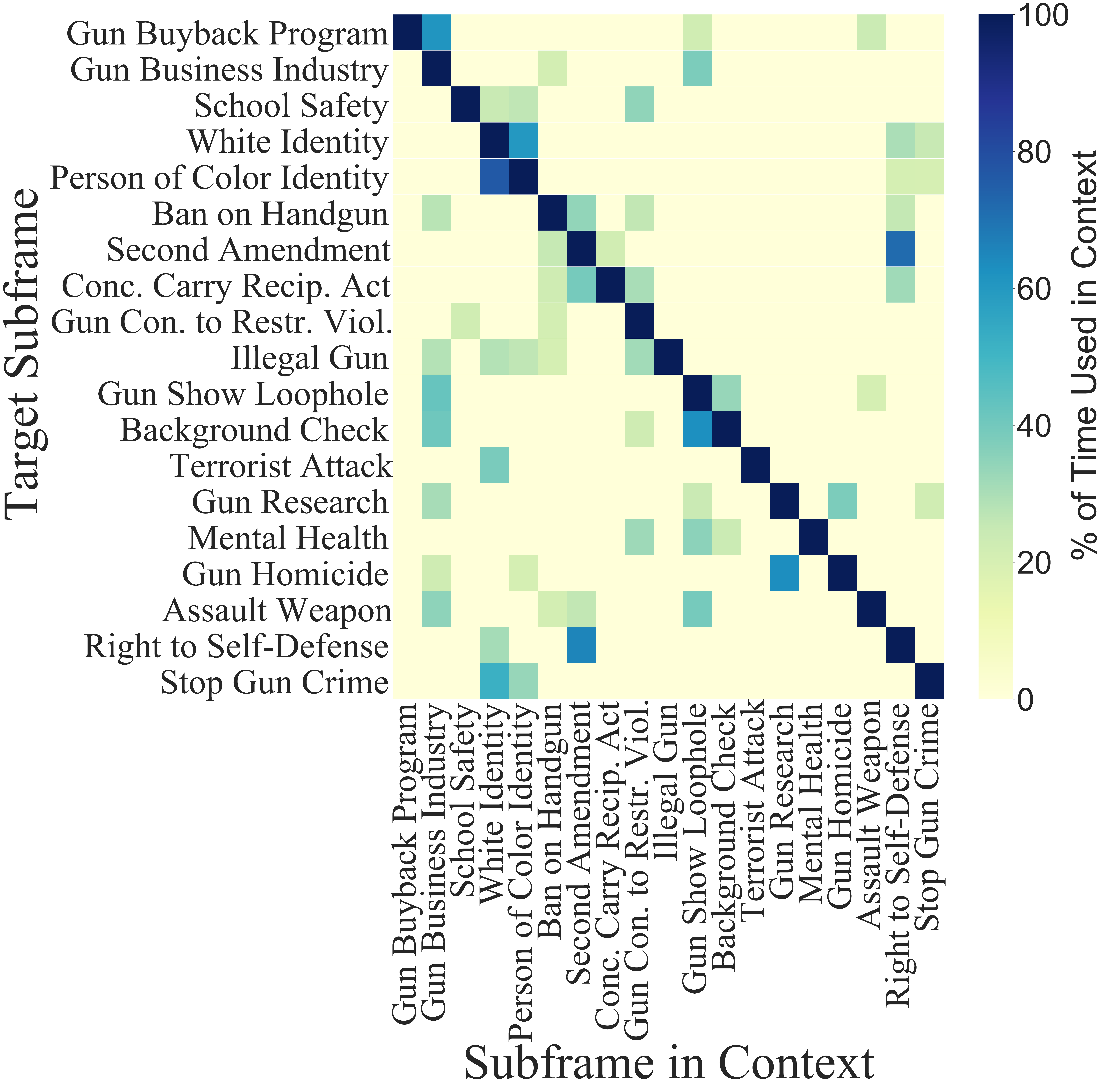}\\
  \caption{Gun Control}
  \end{subfigure}
  \caption{Heatmaps showing subframes used in the context of a subframe on the topics Immigration and Gun Control. The left images are for left biased news articles and the right ones are for right biased ones. Subframes used less than $20\%$ of the time in context are rounded down to zero for a cleaner representation purpose.}
  \label{fig:heatmaps-appendix}
\end{figure*}

The heatmaps showing usage of a subframe in the context of other subframes for the topics immigration and gun control are showed in Figure \ref{fig:heatmaps-appendix}.

\section{Polarization in Frame and Subframe Usage}
\label{Appendix:polarization}
\begin{figure}[ht]
    \centering
    \begin{subfigure}[t]{0.23\textwidth}
        \centering
        \includegraphics[width=\textwidth]{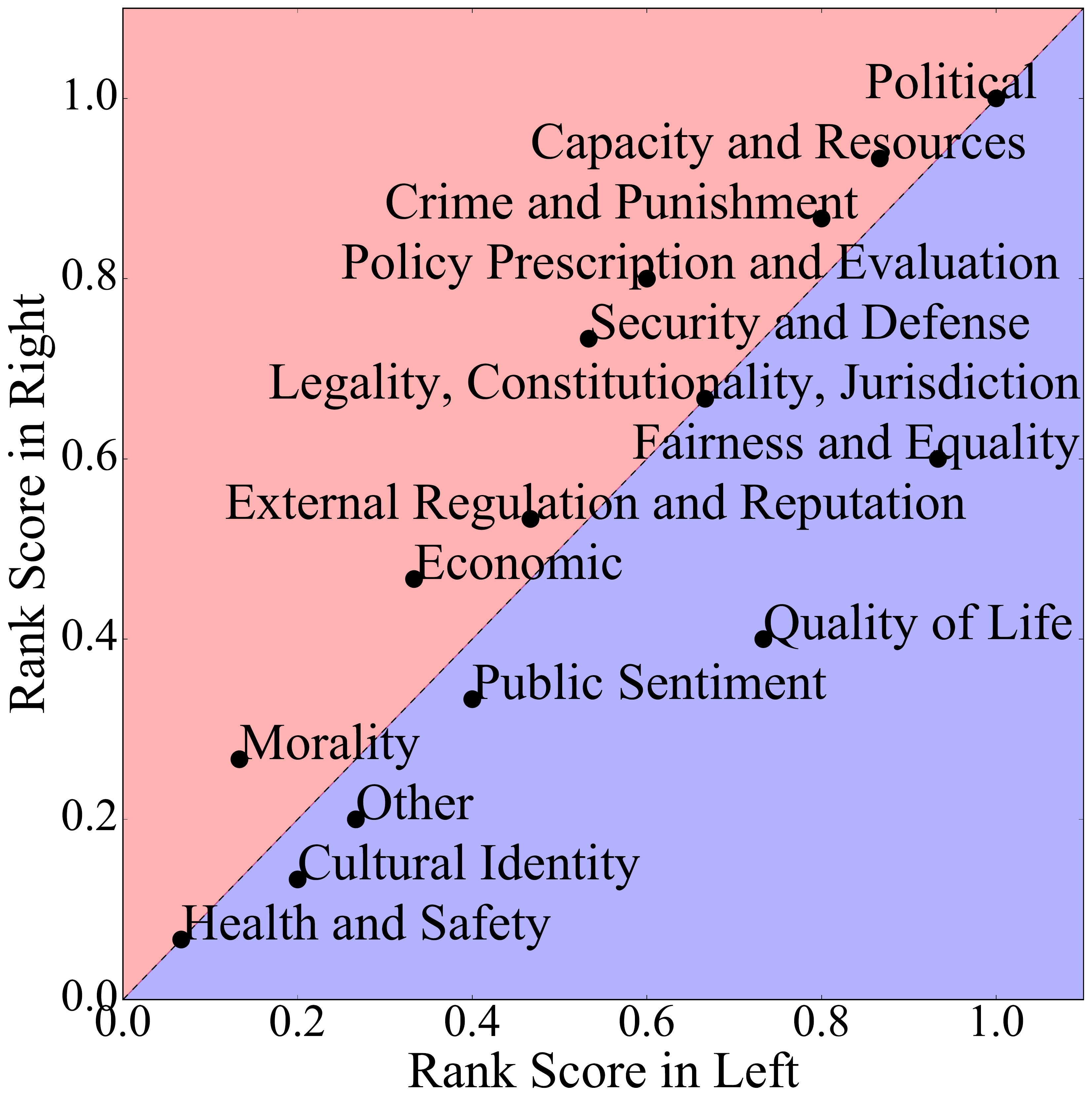} 
        \caption{Frame}
    \end{subfigure}%
    ~ 
    \begin{subfigure}[t]{0.23\textwidth}
        \centering
        \includegraphics[width=\textwidth]{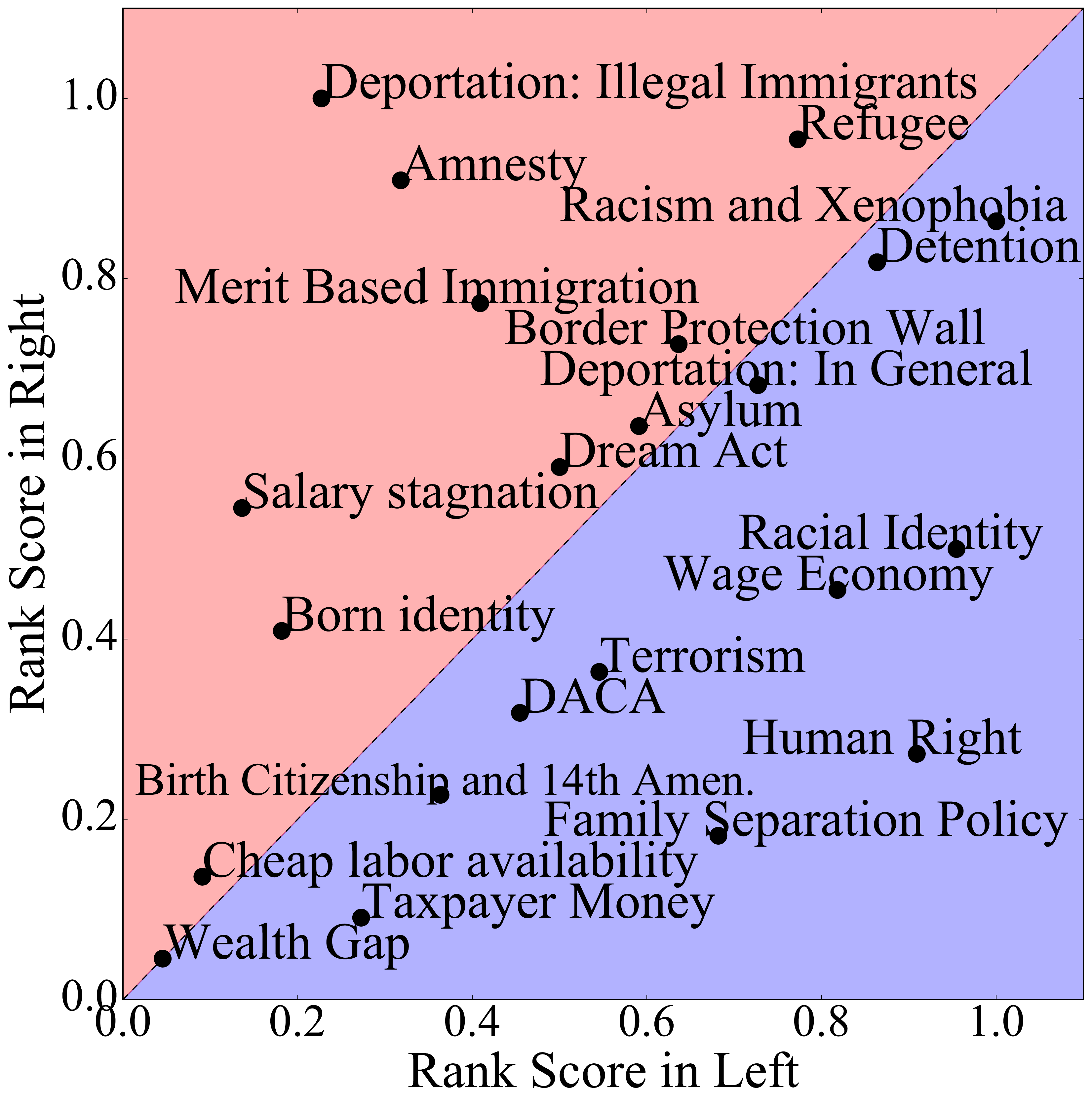}
        \caption{Subframe}
    \end{subfigure}
    \caption{Polarization in usage of frames and subframes. The ranking scores are obtained by taking the normalized rank of the frames and subframes where the highest ranked instance get a score of 1. The rankings are over all news articles in the topic Immigration.}
    \label{fig:polarization-immigration-appendix}
\end{figure}

\begin{figure}[ht]
    \centering
    \begin{subfigure}[t]{0.23\textwidth}
        \centering
        \includegraphics[width=\textwidth]{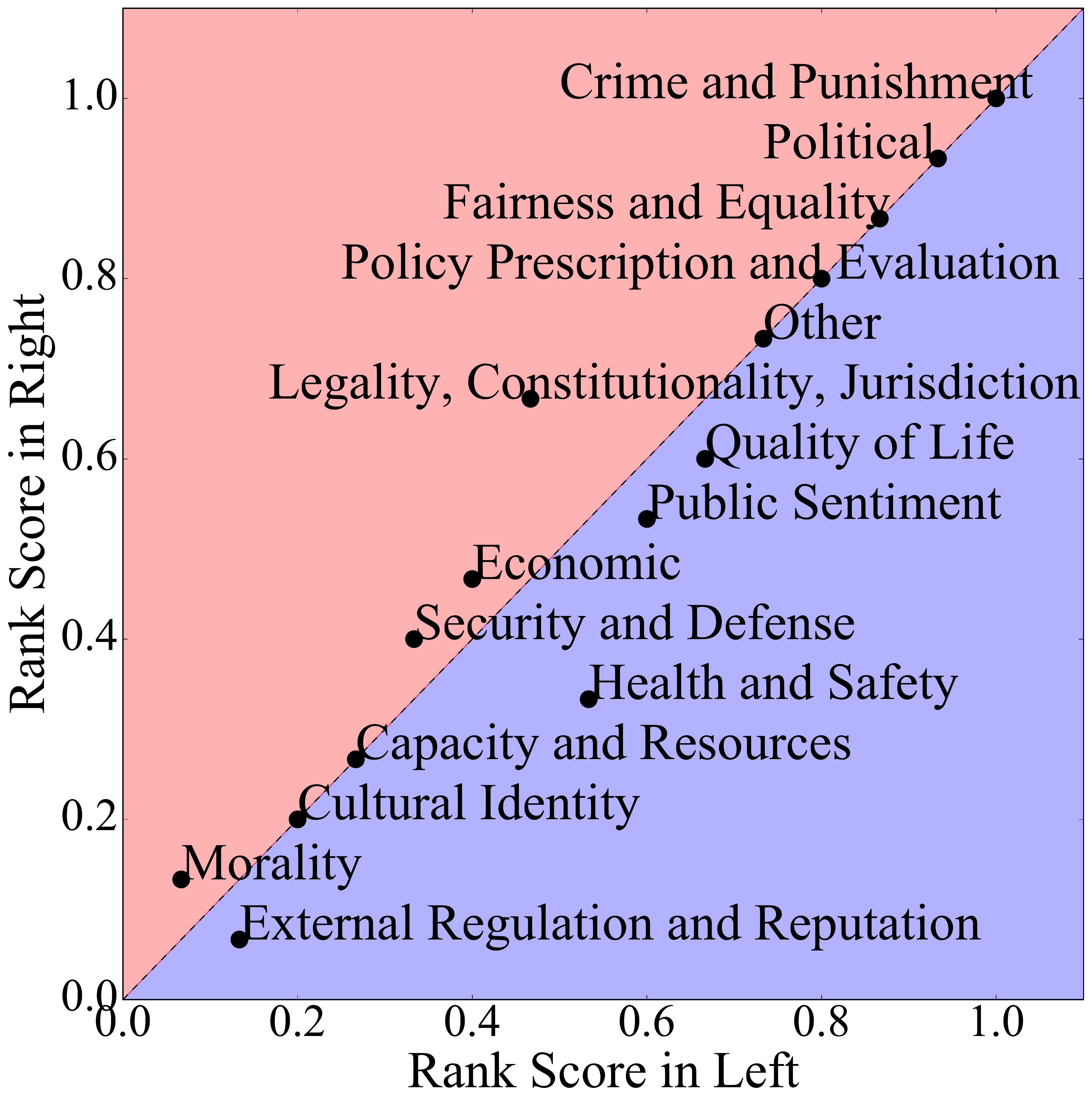}
        \caption{Frame}
    \end{subfigure}%
    ~ 
    \begin{subfigure}[t]{0.23\textwidth}
        \centering
        \includegraphics[width=\textwidth]{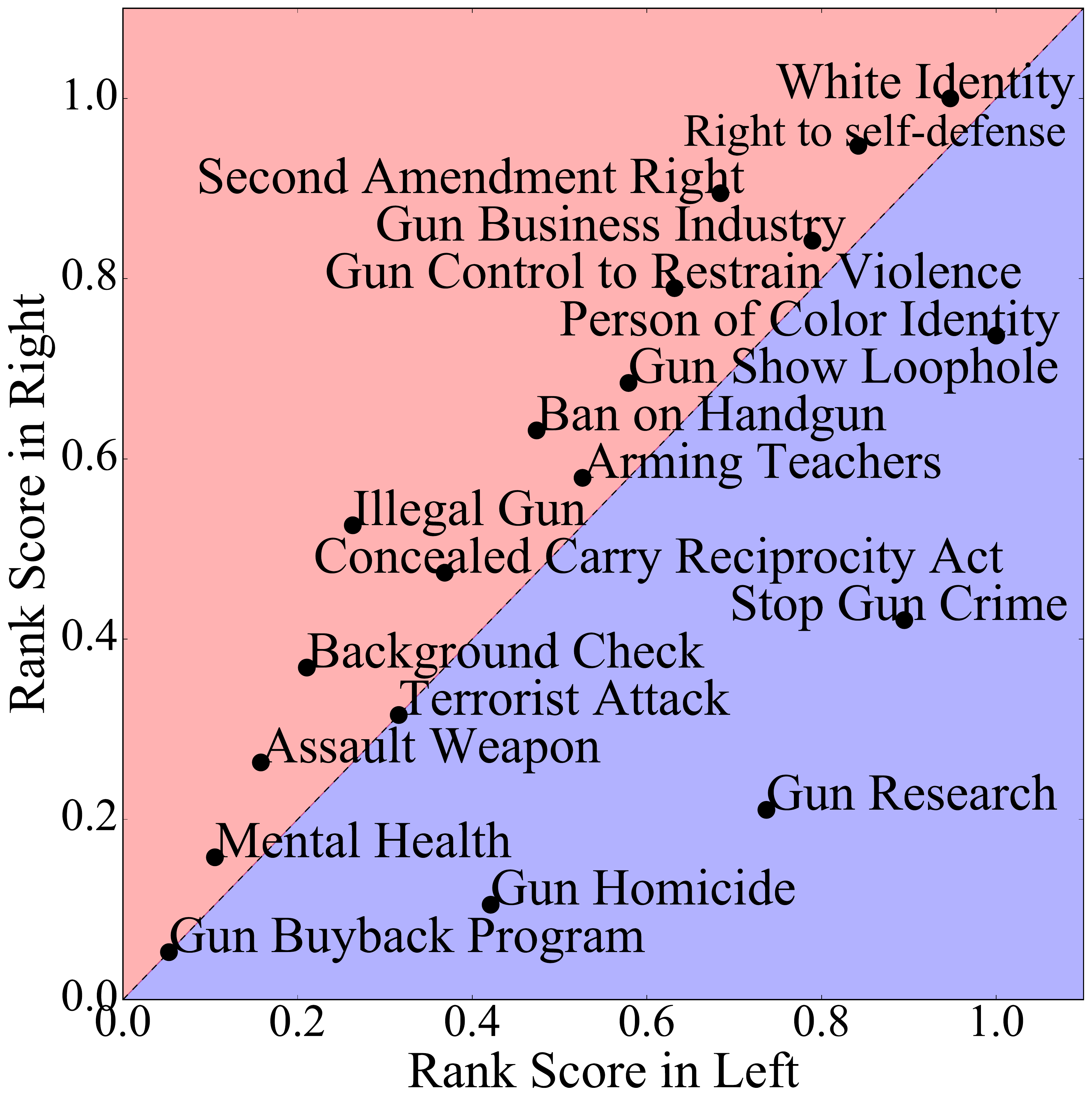} 
        \caption{Subframe}
    \end{subfigure}
    \caption{Polarization in usage of frames and subframes. The ranking scores are obtained by taking the normalized rank of the frames and subframes where the highest ranked instance get a score of 1. The rankings are over all news articles in the topic Gun Control.}
    \label{fig:polarization-gun-appendix}
\end{figure}

Figure \ref{fig:polarization-immigration-appendix} and \ref{fig:polarization-gun-appendix} shows the polarization in usage of frames and subframes by each party for the topics Immigration and Gun Control respectively.

\section{Reproducibility}
\label{Appendix:reproducibility}
\paragraph{Machine Used} We used a Nvidia GeForce GTX 1080 Ti, 12 GB memory GPU in a machine with Intel(R) 12 Core(TM) i7-8700 CPU @ 3.20GHz and a RAM of 64GB to run all the experiments.

\paragraph{Libraries Used} For implementation of HLSTM, BERT and the embedding learning we used PyTorch. To implement the guided LDA we used Python guidedlda library.

\paragraph{Guided LDA Hyper-parameters} We ran the guided LDA models for $100$ iterations with a random state of $7$. This threshold was set by looking at the log-likelihood of the model. For all of the topics at $100$th iteration the log-likelihood became stable. We used seed confidence of $1$ each time which means the seed n-grams had $100\%$ prior probability of being in the corresponding topics i.e. subframes.

\paragraph{Text Classification Baseline} For the text classification baseline, HLSTM, we used validation accuracy as a stopping criteria. If the validation accuracy didn't increase for 10 epochs we stopped training. $10\%$ of the news articles from the training set was used as validation set. This training took on average $10$ minutes for each fold for all of the topics. In BERT training using a batch size of $4$ and learning rate of $5e^{-5}$ yielded the best performance. 

\paragraph{Joint learning of paragraphs and subframe label embeddings} While initializing the embeddings randomly we used a random seed of $1234$. Each learning epoch took on average $67$ seconds for Abortion and Gun Control and $122$ seconds for $Immigration$. We trained the model for at most $100$ epochs or stopped learning if the embedding learning loss didn't decrease for $10$ consecutive epochs.

\end{document}